\documentclass[10pt,journal,compsoc]{IEEEtran}

\ifCLASSOPTIONcompsoc
  \usepackage[nocompress]{cite}
\else
  \usepackage{cite}
\fi

\usepackage{graphicx}
\usepackage{amsmath}
\usepackage{amssymb}
\usepackage{booktabs}
\usepackage{multirow}
\usepackage{epstopdf}
\usepackage{subfig}
\usepackage{color}
\usepackage{url}

\ifCLASSINFOpdf

\else

\fi

\begin{document}

\title{Rich Action-semantic Consistent Knowledge for Early Action Prediction}

\author{Xiaoli~Liu,
        Jianqin~Yin*,
        Di~Guo,
        and~Huaping~Liu
\IEEEcompsocitemizethanks{\IEEEcompsocthanksitem Xiaoli Liu, Jianqin Yin and Di~Guo are the School of
Artificial Intelligence of Beijing University of Posts and Telecommunications,
No.10 Xitucheng Road, Haidian District, Beijing 100876, China.\protect\\
* Jianqin Yin is the corresponding author. \protect\\
E-mail: Liuxiaoli134@bupt.edu.cn, jqyin@bupt.edu.cn
\IEEEcompsocthanksitem Huaping Liu is with the Department of Computer Science and Technology
of Tsinghua University, Beijing 100084, China.  \protect\\
}
}


\IEEEtitleabstractindextext{%
\begin{abstract}
Early action prediction (EAP) aims to recognize human actions from a part of action execution in ongoing videos, which is an important task for many practical applications. Most prior works treat partial or full videos as a whole, ignoring rich action knowledge hidden in videos, i.e., semantic consistencies among different partial videos. In contrast, we partition original partial or full videos to form a new series of partial videos and mine the Action-Semantic Consistent Knowledge (ASCK) among these new partial videos evolving in arbitrary progress levels. Moreover, a novel Rich Action-semantic Consistent Knowledge network (RACK) under the teacher-student framework is proposed for EAP. Firstly, we use a two-stream pre-trained model to extract features of videos. Secondly, we treat the RGB or flow features of the partial videos as nodes and their action semantic consistencies as edges. Next, we build a bi-directional semantic graph for the teacher network and a single-directional semantic graph for the student network to model rich ASCK among partial videos. The MSE and MMD losses are incorporated as our distillation loss to enrich the ASCK of partial videos from the teacher to the student network. Finally, we obtain the final prediction by summering the logits of different subnetworks and applying a softmax layer. Extensive experiments and ablative studies have been conducted, demonstrating the effectiveness of modeling rich ASCK for EAP. With the proposed RACK, we have achieved state-of-the-art performance on three benchmarks. The code is available at \url{https://github.com/lily2lab/RACK.git}.
\end{abstract}

}

\maketitle

\IEEEdisplaynontitleabstractindextext
\IEEEpeerreviewmaketitle

\section{Introduction} \label{sec:intro}
\begin{figure}[!t]
\begin{center}
\includegraphics[width=0.95\linewidth, trim=35mm 1mm 45mm 2mm,clip=true]{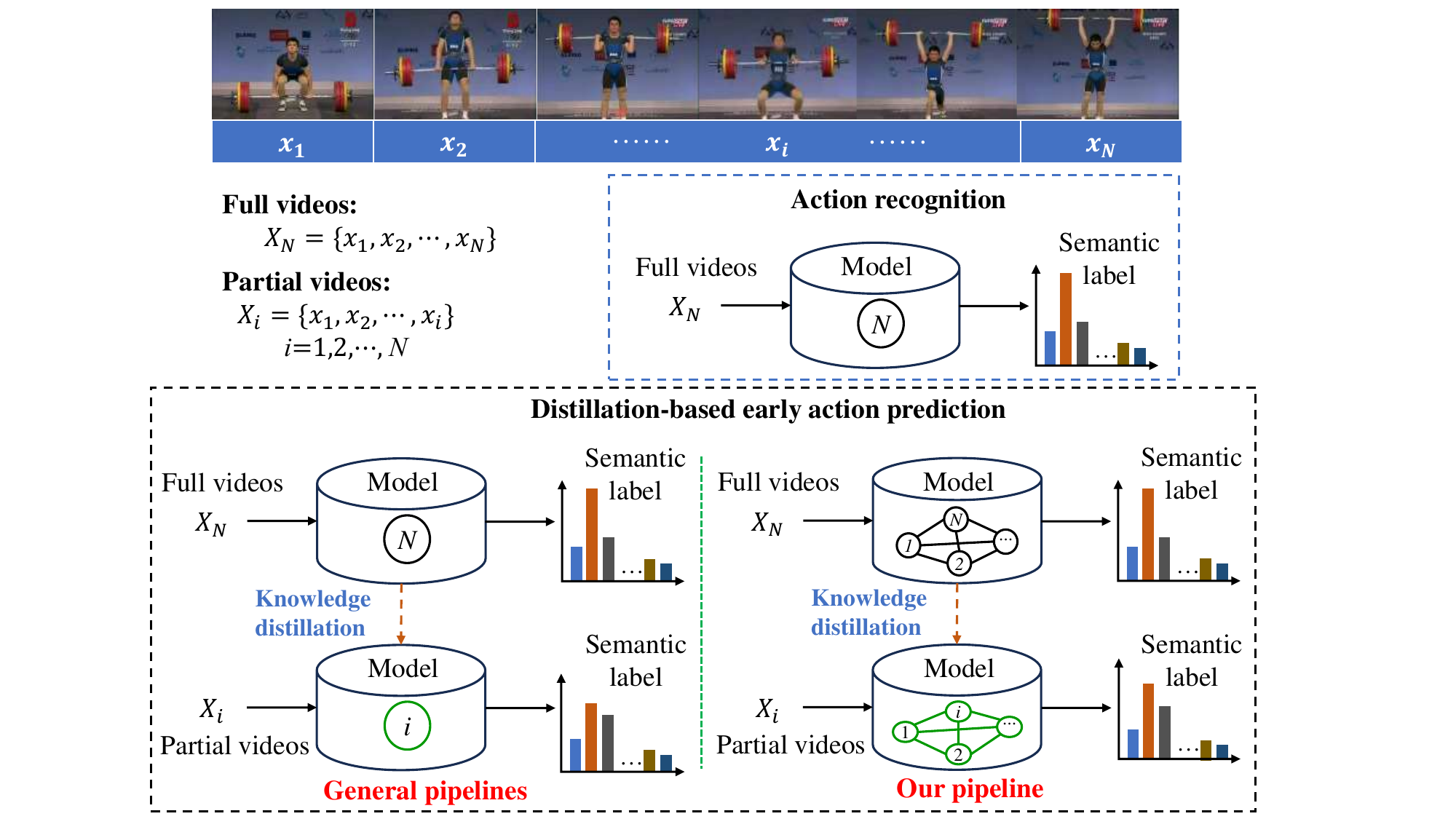}
\end{center}
\caption{Action recognition $vs.$ early action prediction (EAP). {The top one illustrates some concepts, such as partial and full videos. The middle one displays the general pipeline of action recognition. The bottom-left one depicts the general distillation pipelines for EAP, and the bottom-right one showcases our RACK. The numbers in circles indicate the progress levels of partial videos. $N$ is the number of progress levels.}
}
\label{fig:EAR1}
\end{figure}

\IEEEPARstart{E}{arly} action prediction (EAP) is an important task in a series of applications ranging from intelligent surveillance and self-driving to human-robot interaction \cite{zhao2021review,kong2018human}. Unlike traditional action recognition tasks {\cite{yang2016discriminative,yang2016latent}}, as shown in Fig. \ref{fig:EAR1}, the EAP aims to predict the semantic label of ongoing actions very early before the action is completely executed. Due to the limited observations, the action semantics of partial videos are ambiguous, especially at the very early stages. According to the definitions of EAP, the same action execution with different progress levels of videos shares the same action semantics. {However, due to variations in the completion degree of different partial videos, partial videos with different progress levels exhibit diverse feature distributions, even for the same action executions. This leads to significant intra-class variations.} As a result, EAP is particularly challenging, especially in predicting action semantics very early.

Ideally, it is expected that the same semantics of actions should have similar feature distributions. Therefore, the key to EAP is to learn similar distributions of partial videos of the same action executions \cite{wang2019progressive,kong2017deep}.
Similar distributions of partial videos can be measured by correlations of their features, which is called action-semantic consistent modeling. Through pairwise action-semantic consistent modeling shown in Fig. \ref{fig:EAR1}, we may learn similar distributions and their action-semantic consistencies, and we name it Action-Semantic Consistent Knowledge (ASCK). Therefore, modeling ASCK among arbitrary partial videos is an intuitive way to learn the similar distributions of their features for alleviating the large intra-class variances of partial videos.

Many prior works have been proposed for EAP, mainly including recognition-based methods and distillation-based methods. The recognition-based methods directly predict semantic labels of partial videos which contain limited observations with incomplete action executions \cite{chen2018part,sun2019relational,liu2018ssnet}. For example,
Sun et al. \cite{sun2019relational} proposed a discriminative relational recurrent network (DR2N) and modeled spatio-temporal interactions between actors of several historical frames for predicting future actions.
Due to limited observations, the partial video shares ambiguous semantics, which inevitably leads to poor performance by the recognition-based methods.

The distillation-based methods transfer additional action knowledge from full videos for the missing information of partial videos, which can effectively reduce the ambiguous semantics for predictions \cite{gammulle2019predicting,cai2019action,wang2019progressive}. However, as shown in Fig. \ref{fig:EAR1}, most existing works treat the partial or full videos as a whole, ignoring the action-semantic knowledge among partial videos with lower progress levels hidden in the original video. Therefore, these models cannot capture richer ASCK for accurate predictions. By contrast, as shown in Fig. \ref{fig:EAR1}, we partition original partial or full videos into a series of new partial videos with {lower} progress levels. We model rich ASCK among these new partial videos for predictions.

In this paper, we propose a novel Rich Action-semantic Consistent Knowledge network (RACK) {for EAP, utilizing both the graph neural network and teacher-student framework}. Specifically, we build the teacher network using a bi-directional fully connected graph with partial videos as nodes. The teacher network distills ASCK among partial videos from lower to higher progress levels and higher to lower progress levels. The student network is built with a single-directional fully connected graph since future information is unavailable during the testing phase. The student network distills ASCK among partial videos from lower to higher progress levels. We further transfer the ASCK from full to partial videos via layer-wise distillation loss between teacher and student networks. With these elegant designs, the student network also distills ASCK from higher to lower progress levels for predictions. Moreover, we extract two-stream features as the representation of partial or full videos, including RGB and flow features. We obtain the logits of RGB or flow modalities by feeding the two-stream features to RACK, respectively. Finally, we summarize these logits and obtain the predictions via a softmax layer.
In contrast to \cite{wang2019progressive}, we capture richer ASCK among partial videos with arbitrary progress levels instead of adjacent partial videos.
We have achieved a new state-of-the-art performance on three benchmarks (i.e., UCF101, HMDB51, Something Something V2), showing the effectiveness of modeling rich ASCK.

Our main contributions are three folds.
\begin{itemize}
\item To the best of our knowledge, we are the first to explore rich ASCK among arbitrary partial videos, which helps to learn similar feature distributions of partial videos with the same action execution for accurate predictions.
\item  {We propose a novel RACK model for EAP, utilizing the graph reasoning and knowledge distillation frameworks. The proposed method explores rich action knowledge for accurate predictions via graph-based knowledge reasoning and knowledge distillation.}
\item {The experimental results demonstrate our state-of-the-art performance on three datasets, highlighting the effectiveness of our method. Furthermore, extensive ablative analysis has revealed that modeling rich ASCK is crucial in improving predictions.}
\end{itemize}

\section{Related work}
\subsection{Early Action prediction}
\noindent Most existing works follow the pipeline of action recognition by directly recognizing the semantic label from partial observations \cite{hu2018early,lai2017global,ke2019learning}. {For example,} Kong et al. \cite{kong2018action} proposed a men-LSTM model to remember the hard-to-predict samples, which forced their model to learn a more complex classified boundary for accurate predictions. {However, due to the inherent weakness of LSTM on long-term memorization, it was difficult to remember all hard-to-predict samples.}
Similarly, Li et al. \cite{li2020hard} modeled relationships between similar action pairs and dynamically marked the hard-predicted similar pairs for 3D skeleton-based EAP. The author used an adversarial learning scheme to generate hard instance features, which was difficult to train.
The above models mainly focused on high-level semantics, which ignored low-level movement for actions. Therefore,
Wu et al. \cite{wu2021spatial} and Yao et al. \cite{yao2018multiple} modeled {spatial} interaction relationships and their evolutions for EAP. Li et al. \cite{li2012modeling,li2014prediction} mined sequential temporal patterns via causality of continuing action units, context cues, and action predictability for the long-duration action prediction.
The recognition-based methods easily suffered from poor performance due to the limited observations and large intra-class variations among partial videos with different progress levels.

Two standard pipelines were proposed to alleviate the semantic ambiguity of the limited observations, including future feature reconstruction of future videos \cite{shi2018action,xu2015activity,ng2020forecasting} and knowledge distillation from full videos \cite{kong2017deep,cai2019action,wang2019progressive}.
In terms of future feature reconstruction, \cite{fernando2021anticipating,pang2019dbdnet,guan2020generative} jointly learned future motions and relationships between the observed and future information for early predictions. Kong et al. \cite{kong2017deep} exploited abundant sequential context features {utilizing} full videos to enrich and reconstruct the features of partial videos.
Regarding knowledge distillation, Cai et al. \cite{cai2019action} learned {future} embeddings by transferring the knowledge between partial and full videos. {To further enrich the knowledge distillation,} Wang et al. \cite{wang2019progressive} learned progressive action knowledge between adjacent partial videos but ignored semantic consistencies of partial videos between lower and higher progress levels, {also} suffering from the limited ability to learn rich action knowledge.
By contrast, we model rich ASCK among arbitrary partial videos via densely graph relationship modeling.

\subsection{Graph neural network (GNN)}
\noindent Traditional GNNs modeled point-level relationships for non-Euclidean data, such as point cloud \cite{liu2019learning} and human skeleton data \cite{GNNvs}. For example, many researchers treated human body joints as nodes and their relationships as edges for skeleton-based action recognition \cite{stgcn,shiftgcn} or human motion prediction \cite{mao2019learning}. Recently, GNN has been extended to a series of visual tasks \cite{chen2019graph,wang2018videos,yuan2017temporal}{, including object-level relationships \cite{wang2018videos,qi2018learning}, frame-level relationships \cite{li2021video,lu2021segmenting}, proposal-level relationships \cite{zeng2019graph,bai2020boundary}, snippet-level relationships \cite{zhao2021video,xu2020g}, etc. For example, \cite{qi2018learning} set the detected humans or objects as nodes and their relationships as edges to model the object-level relationships for human-object interaction (HOI) detection or recognition tasks. \cite{li2021video,lu2021segmenting} modeled the frame-level relationships with frames as the nodes and their relationships as edges, capturing spatio-temporal clues for recognition and segmentation tasks. \cite{zeng2019graph} modeled proposal-level relationships by exploring relationships between proposals and action contents.}
\cite{zhao2021video,xu2020g} modeled snippet-level relationships by taking video snippets as nodes and their relationships as edges to exploit the correlations among video snippets. {\cite{zheng2019reasoning} explored the underlying dialog structure among visual caption and question \& answer pairs for reasoning visual dialogs.}

However, prior methods modeled spatio-temporal relationships of text pairs, objects, snippets, or frames instead of the action semantics in videos. Due to the domain gaps of different tasks, these methods can not be directly adopted for EAP. In contrast to prior works, we explore video-level relationships by treating the whole partial or full videos as nodes and their semantic consistencies as edges, modeling rich action-semantic consistencies between different videos.

\subsection{{Semantic consistent knowledge}}
\noindent Semantic consistent knowledge has been explored in many visual tasks. For example, Xie et al. \cite{xie2020multi} and Chen et al. \cite{chen2019cross} modeled semantic consistency between image and text modalities for cross-model retrieval. Jia et al. \cite{jia2021spatial} modeled spatial and semantic consistency for pedestrian attribute recognition. Gao et al. \cite{gao2017video} explored the semantics between words and visual content for video captioning. Different from these tasks, we focus on exploring action semantic consistency among different partial videos.

Recently, many models were proposed based on knowledge distillations under the teacher-student framework
\cite{kdsurvey,KD}. For example,
Cai et al. \cite{cai2019action} and Wang et al. \cite{wang2019progressive} introduced the knowledge distillation framework for EAP. These models tried to distill action knowledge from full videos by the teacher network and transferred it to the partial videos by the student network. However, Cai et al. \cite{cai2019action} ignored semantic correlations among different partial or full videos, and Wang et al. \cite{wang2019progressive} modeled correlations between adjacent partial videos. In contrast to \cite{cai2019action} and \cite{wang2019progressive}, we consider rich semantic correlations between any two partial or full videos and model their ASCK.

\section{Methodology}
\subsection{Problem formulation} \label{sec:3.1}
\noindent Consistent with the existing works \cite{kong2017deep,wang2019progressive,zhao2021review}, we assume that a video contains the action with complete executions. Given a video ${\bf{\emph X}}$, we uniformly partition it into $N$ segments as $\{{\bf{\emph x}}_1,{\bf{\emph x}}_2,\cdots,{\bf{\emph x}}_N \}$. The first $n$ segments form a partial video ${{\bf{\emph X}}_n} = [{{\bf{\emph x}}_1},{{\bf{\emph x}}_2}, \cdots ,{{\bf{\emph x}}_n}]$ with a progress level of $n$, and its observation ratio is defined as ${n \mathord{\left/
 {\vphantom {n N}} \right.
 \kern-\nulldelimiterspace} N}$, which represents the action completion degree. Early action prediction (EAP) can be formulated as equation \ref{eq1}.

\begin{equation}
{c_n} = f({{\bf{\emph X}}_n},n)
  \label{eq1}
\end{equation}
where $f( \cdot )$ is a mapping that transforms a partial video, ${\bf{\emph X}}_n$ into a semantic label $c_n$, $n$ denotes the progress level of partial video ${{\bf{\emph X}}_n}$, and $n<N$ for the EAP.

As shown in Fig. \ref{fig:EAR1}, the distillation-based models utilize full videos to obtain additional information for partial videos with incomplete action execution, effectively assisting predictions. Following this pipeline, we build a RACK network under the teacher-student framework to parameterize $f( \cdot )$ in section \ref{sec:method}. In contrast to prior works, we model rich ASCK among partial videos with arbitrary progress levels, which assists in learning missing information and reducing the semantic ambiguities of partial videos.

\subsection{Early action prediction}  \label{sec:method}
\begin{figure*}[!t]
  \centering
  \includegraphics[width=0.9\linewidth]{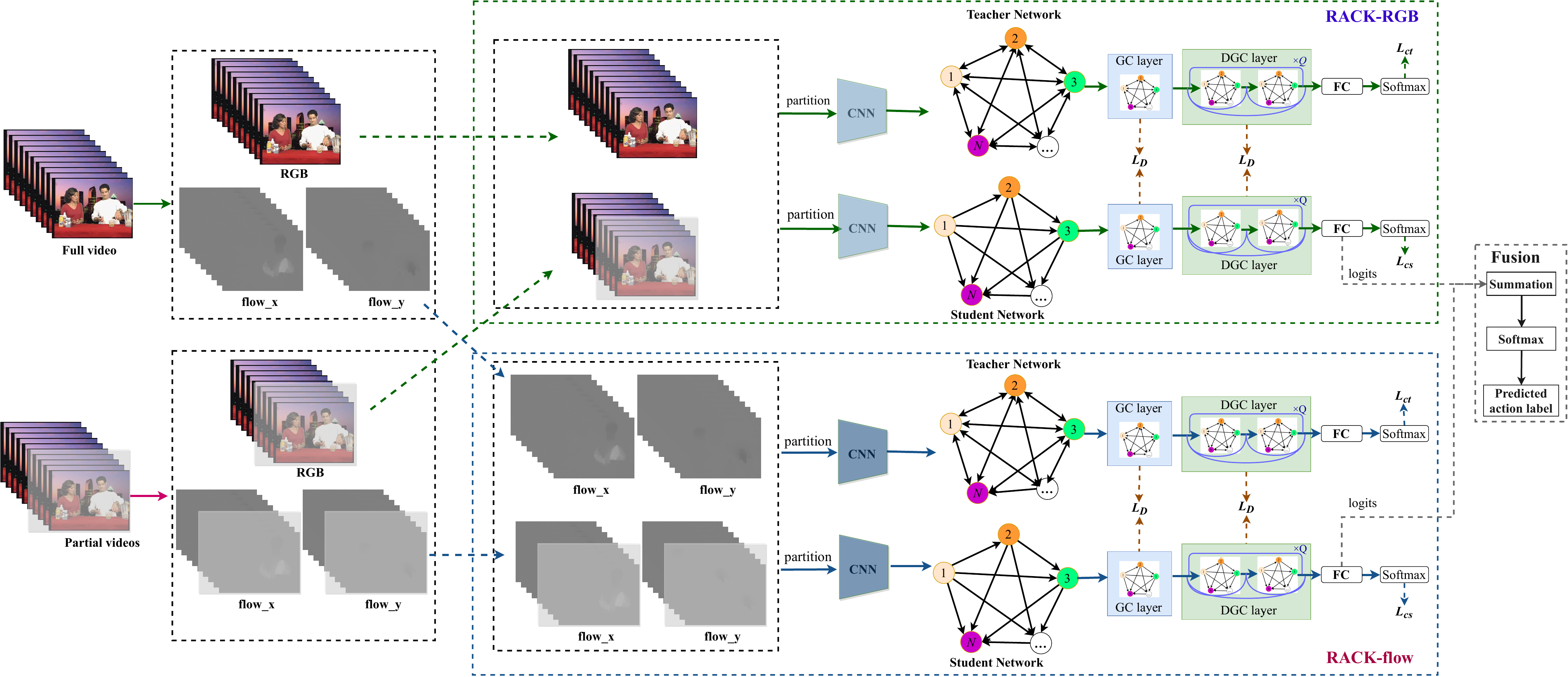}
   \caption{Overall framework. A novel RACK network is built under the teacher-student framework for early action prediction.}
   \label{fig:overallframework}
\end{figure*}

\noindent Fig. \ref{fig:overallframework} shows our RACK network under the two-stream framework for EAP, which respectively takes RGB and optical flow images as inputs, including RACK-RGB and RACK-flow. Each stream mainly includes the teacher network, student network, and loss. The teacher or student network learns ASCK of partial videos hidden in the original videos, respectively. The ASCK is transferred from the teacher to the student via the distillation loss, which enriches the ASCK of partial videos for early predictions. Besides, we utilize the advantages of relational modeling of GNNs to model ASCK among partial videos with arbitrary progress levels. Furthermore, graph generation is the key to building teacher and student networks. In the following subsections, we show graph generation, two backbone layers, the network structure, and our distillation loss.

\subsubsection{Graph generation} \label{sec:3.3}
\noindent Consistent with prior works \cite{hu2018early,cai2019action,wang2019progressive}, we assume that full videos contain complete action execution. That is to say, the full videos implicitly include partial videos with arbitrary progress levels. Similarly, current partial videos contain other partial videos with lower progress levels. Based on the observation above, the original partial or full videos can be partitioned into a new series of partial videos with lower progress levels. In this way, we can explicitly model ASCK of partial videos with different progress levels for predictions.

Generally, the teacher network is applied to full videos, and the student network is applied to partial videos with different progress levels.
Therefore, we build a bi-directional fully connected graph for the teacher network, and we build the other single-directional fully connected graph for the student network since future information is not available during the testing phase. In this way, the teacher network models ASCK among partial videos with arbitrary progress levels. The student network models the ASCK among partial videos from lower to higher progress levels. As a result, we build two graphs for the teacher and student networks as follows.

\begin{figure*}[!t]
  \centering
  \includegraphics[width=0.95\linewidth]{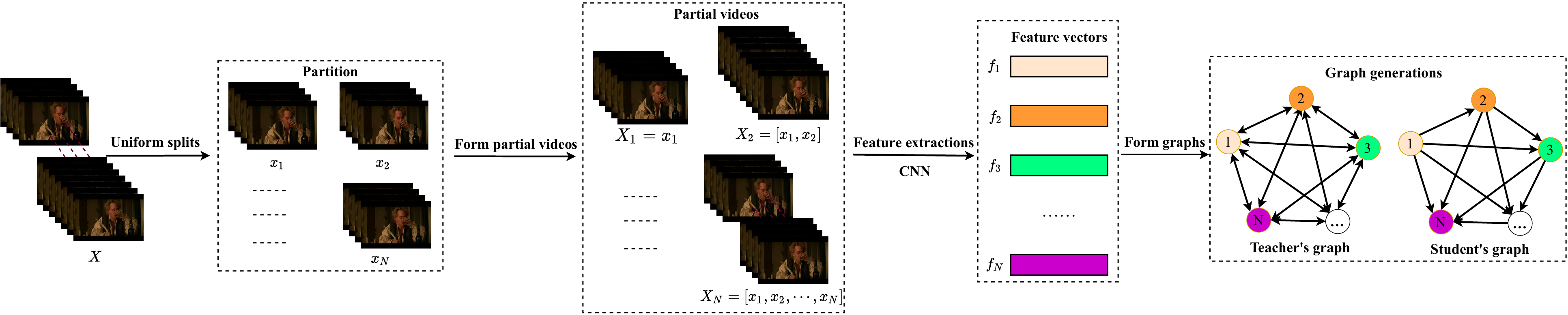}
   \caption{Graph generation. Given a video {\bf{\emph X}}, we first uniformly split it into $N$ segments $\{{\bf{\emph x}}{_1},{{\bf{\emph x}}_2}, \cdots ,{{\bf{\emph x}}_N}\}$. Then, the partial video ${\bf{\emph X}}_n$ with a progress level of $n$ can be formed by the first $n$ segments via concatenation along the temporal dimension. Next, taking the RGB modality as an example, we calculate the frames of partial videos \cite{tsn} and apply the CNN model to extract their spatio-temporal features using RGB images, respectively. ${\bf{\emph f}}_n$ denotes the extracted features of the partial video ${\bf{\emph X}}_n$ ($n = 1, 2, \cdots, N$). Finally, the graphs are built for the teacher and student networks with the features of partial videos as nodes and their semantic consistencies as edges. }
   \label{fig:graphgeneration}
\end{figure*}

As shown in Fig. \ref{fig:graphgeneration}, firstly, we partition the original video ${\bf{\emph X}}$ to obtain new partial videos with different progress levels as defined in section \ref{sec:3.1}. Then, we extract two-stream features for representing the partial videos, including RGB features and flow features. Finally, two graphs are defined for the teacher and student networks, using RGB or flow features of partial videos as nodes and their semantic consistencies as edges.

Feature extraction. We extract features of partial videos using different backbones, including R(2+1)D \cite{tran2018closer}, 3D ResNet \cite{hara2018can} and BN-Inception \cite{tsn}. Moreover,
two-stream networks have succeeded in various visual tasks {\cite{tsn,xie2019semantic}}. In this paper, we also follow this pipeline. As described in \cite{yin2018one}, we also use a pre-trained BN-Inception network \cite{BNInception} on the Kinetics-400 dataset \cite{kay2017kinetics} to extract the features using RGB and optical flow image sequences of partial or full videos.
Besides, we also use the pertained models on the Kinetics-700 implemented by \cite{hara2018can} to extract features of R(2+1)D and 3D ResNet \footnote{{\url{https://github.com/kenshohara/3D-ResNets-PyTorch}}}.
Taking the optical flow image sequence as an example, we use a sliding window with a size of $P$ and an overlap of $O$ to collect a batch of optical flow images. The two-directional optical flow images are stacked and resized as a tensor with a size of $2*P \times H_i \times W_i$, where $H_i$ and $W_i$ denote the height and width of the images. These tensors are set as the input of the BN-Inception network, and the output of the last fully connected (FC) layer (generally before the softmax layer) is set as the features of stacked images. Then, we obtain a sequence of feature vectors with a size of $1 \times D$ for each partial or full video, where $D$ is determined by the output dimensions of the last FC layers. Furthermore, we apply an average pooling to these feature vectors and obtain the final feature representation with a size of $1 \times D$ for a partial or full video.
Similarly, a window of RGB images is stacked and resized as a tensor with a size of $P \times H_i \times W_i$, and the features are obtained by setting these tensors as the input of the backbones, respectively.
Finally, we extract RGB and flow features of partial or full videos.

\subsubsection{Graph convolutional (GC) layer} \label{sec:3.4}
\noindent GNNs have shown a solid ability to model various relationships \cite{GNNvs}.
To utilize these advantages for modeling action semantic consistencies among partial or full videos, we define a graph convolutional network layer as follows by equation \ref{eq2}.

\begin{equation}
  \begin{array}{c}
{{\bf{\emph F}}_{l + 1}} = GC({{\bf{\emph F}}_l})\\
 = {{\bf{\emph A}}_l}{{\bf{\emph F}}_l}{{\bf{\emph W}}_l}
\end{array}
  \label{eq2}
\end{equation}
where $GC\left(  \cdot  \right)$ denotes the defined graph convolutional layer, ${\bf{\emph F}}_l$ denotes a tensor with a shape of $N \times D_l$ at the $l$-th layer. {We denote $f_i={{\bf{\emph F}}_l}(i,:)$ as the $i$-th row of ${{\bf{\emph F}}_l}$, which represents the features of a partial video with a progress level of $i$. In the paper, we initialize $f_i$ using the features of partial videos as described in section \ref{sec:3.3}.} ${\bf{\emph W}}_l$ is a learnable weight with a shape of $D_l \times D_{l+1}$, and $D_l$ is the dimension of the $l$-th layer. ${\bf{\emph A}}_l$ denotes the adjacent matrix with a shape of $N \times N$, and $N$ denotes the number of progress levels of partial videos.

Instead of simply initializing the adjacent matrix with random weights as done in prior works\cite{mao2019learning}, we add a constrained matrix to the adjacent matrix by an element-wise multiplication. The constrained matrix is calculated by the cosine similarity between two partial or full videos, encouraging the network to maximize their similarities and model their semantic consistencies. Therefore, our adjacent matrix ${\bf{\emph A}}_l$ can be formulated as follows by equation \ref{eq3}.

\begin{equation}
  {{\bf{\emph A}}_l} = {\bf{\emph M}} \odot {\bf{\emph A}}' \odot {\bf{\emph S}}
  \label{eq3}
\end{equation}
where ${\bf{\emph M}}$ is a binary mask matrix that indicates the indexes of partial videos for distillation. ${\bf{\emph A'}}$ denotes a learnable adjacent matrix as done prior works \cite{mao2019learning}, and ${\bf{\emph S}}$ is a introduced constrained matrix. The shape of ${\bf{\emph M}}$, ${\bf{\emph A'}}$ or ${\bf{\emph S}}$ is $N \times N$, and $\odot$ denotes an element-wise multiplication.

The constrained matrix $S$ is formulated as equation \ref{eq4}.

\begin{equation}
  {s_{ij}} = \frac{{{f_i}f_j^T}}{{{{\left\| {{f_i}} \right\|}_2}{{\left\| {{f_j}} \right\|}_2}}}
  \label{eq4}
\end{equation}
where ${s_{ij}}$ denotes an element of ${\bf{\emph S}}$ at the $i$-th row and $j$-th column, which can be calculated by the cosine similarity of partial videos between {the $i$-th and $j$-th progress level.
$i, j = 1, 2,\cdots, N$. $f_i={{\bf{\emph F}}_l}(i,:)$ denotes the features of a partial video with a progress level of $i$.} ${\bf{\emph f}}^T$ denotes the transpose of vector ${\bf{\emph f}}$.

\subsubsection{Densely graph convolutional block (DGC)} \label{sec:3.5}
\noindent To improve the model's ability, we build a dense graph convolutional block using the predefined GC layer formulated by equation \ref{eq5}.
\begin{equation}
  \begin{array}{l}
{{\bf{\emph F}}_{l + 1}} = DGC({{\bf{\emph F}}_l})\\
 = g(g({{\bf{\emph F}}_l}) + {{\bf{\emph F}}_l}){\rm{ + (}}g({{\bf{\emph F}}_l}) + {{\bf{\emph F}}_l}) + {\bf{\emph F}}_l
\end{array}
  \label{eq5}
\end{equation}
where $g(\cdot)$ denotes a graph convolutional layer $GC(\cdot)$ followed by a batch normalization layer $BN(\cdot)$, a ReLU layer and a dropout layer $\delta (\cdot)$, i.e., $g(*) = \delta (BN(GC(*)))$.

The dense connections enable the network to fuse the low-level features from shallow layers via element-wise summation, enhancing the features of each partial video to some extent.

\subsubsection{Network structure}
\noindent Based on the generated graphs and the predefined backbone layers, we build our RACK network using the teacher-student framework and the two-stream framework for EAP, including RACK-RGB, RACK-flow, and fusion module. As shown in Fig. \ref{fig:overallframework}, the RACK-RGB and RACK-flow share the same structure, which takes RGB and flow features as the inputs, respectively. The fusion module obtains the final predictions via summarization logits of RACK-RGB and RACK-flow subnetworks. The RACK-RGB or RACK-flow contains a teacher network and a student network. Below, we first show the detailed structures of teacher and student networks and the fusion module. Then, we discuss the {ASCK} of the proposed method.

{\bf Teacher network.} We build a bi-directional action-semantic graph for the teacher network as described in Section \ref{sec:3.3}. More specifically, the element values of binary mask matrix $M$ in equation \ref{eq3} are set to 1, i.e. $m_{ij}$=1, ${m_{ij}} \in M$ and $i, j = 1,2,\cdots, N$. In this way, the teacher network models ASCK among any partial or full videos from lower to higher progress levels and higher to lower progress levels.

The detailed structure of the teacher network is shown in Fig. \ref{fig:overallframework}, including a GC layer, $Q$ stacked DGC layers, a fully connected (FC) layer, and a softmax layer, among which the GC and DGC layers are first applied to update the features of nodes and edges of the graph, so as to model the ASCK among different videos with arbitrary progress levels. Then, the FC and softmax layers are applied to obtain the semantic labels of videos.

{\bf Student network.} The student network shares the same structure as the teacher network except for the single-directional action-semantic graph. Therefore, the binary mask matrix $M$ in equation \ref{eq3} is defined by equation \ref{eq6}. In this way, the partial video can not obtain the information from the following partial videos with higher progress levels. The ASCK is modeled among partial videos from lower to higher progress levels. Nonetheless, the distillation loss also transfers the ASCK from higher to lower progress levels through the teacher-student framework.

\begin{equation}
  {m_{ij}} = \left\{ \begin{array}{l}
1,{\rm{ }}i \ge j\\
0,{\rm{ otherwise}}
\end{array} \right.
  \label{eq6}
\end{equation}
where ${m_{ij}} \in M$ and $i, j =1,2,\cdots,N$.

Furthermore, the FC and softmax layers of the student model are shared with that of the teacher network, which jointly trains the classifiers for EAP.

{\bf Fusion.} Given the logits of partial video $X_i$ after FC layers of RACK-RGB and RACK-flow, denoting by $l_{ir}$ and $l_{if}$ ($i=1,2,\cdots,N$), we obtain the final prediction via element-wise summation of the logits and a softmax layer by equation \ref{eq7}.

\begin{equation}
{\widehat y_{i}} = {\rm{softmax}}({l_{ir}} \odot {l_{if}})
  \label{eq7}
\end{equation}
where ${\widehat y_{i}}$ is the predicted label of partial videos. ${l_{ir}}$ and ${l_{if}}$ are logits of the student network from RACK-RGB and RACK-flow, respectively. $\odot$ is an element-wise summation. ${\rm{softmax}}( \cdot )$ is a softmax layer.

{\bf Rich ASCK.} As discussed above, the proposed RACK network models ASCK from two-fold: (1) graph reasoning-based ASCK. Based on the defined action-semantic graphs and graph backbone layers in sections \ref{sec:3.3}, \ref{sec:3.4}, and \ref{sec:3.5}, we model ASCK among different partial videos via the teacher and student networks. Specifically, the bi-directional semantic graph of the teacher network models ASCK among arbitrary partial videos that are partitioned from the full videos. The single-directional semantic graph of the student network models ASCK of partial videos from lower to higher progress levels. With these elegant designs, the teacher network obtains complete ASCK, and for the student network, the partial videos can utilize the ASCK from partial videos with lower progress levels to improve their semantic predictions.
(2) Distillation framework-based ASCK. Under the knowledge distillation framework, the action knowledge of partial videos is further enriched by distilling complete ASCK from the teacher to the student. We model rich ASCK for accurate predictions via graph reasoning and knowledge distillation.

\subsubsection{Loss}
\noindent As shown in Fig. \ref{fig:overallframework}, we first train RACK-RGB and RACK-flow subnetworks separately. Then, we obtain final predictions via the fusion module. The RACK-RGB and RACK-flow share the same structure and loss. {Taking} an example of RACK-RGB, we optimize our model using the loss defined in equation \ref{eq8}, which jointly achieves ASCK distillation and early action-semantic predictions.

\begin{equation}
L = {L_D} +{L_C}
  \label{eq8}
\end{equation}
\begin{equation}
\begin{array}{c}
{L_D} = {L_{MSE}} + {L_{MMD}} \\
= {\sum\nolimits_{l = 1}^{Q+1} {(\left\| {{F_{ls}} - {F_{lt}}} \right\|} _2} + \left\| {{F_{ls}}F_{ls}^T - {F_{lt}}F_{lt}^T} \right\|)
\end{array}
  \label{eq9}
\end{equation}
\begin{equation}
\begin{array}{c}
{L_C} = {L_{CT}} + {L_{CS}}\\
 = \sum\nolimits_{i = 1}^N {(CE({{\widehat y}_{it}},{y_i})}  + CE({\widehat y_{is}},{y_i}))
\end{array}
  \label{eq10}
\end{equation}
where {$Q$ denotes the number of stacked DGC layers. ${F_{l*}}$ denotes the feature map of the student or teacher network at the $l$-th GC or DGC layer.} $L_D$ denotes the layer-wise knowledge distillation loss built with MSE and MMD as done in \cite{wang2019progressive}.
$L_C$ denotes the overall classification loss, including a classification loss of the teacher network $L_{CT}$ and a classification loss of the student network $L_{CS}$. $CE(\cdot)$ denotes the standard cross-entropy loss between the prediction ${\widehat y_{i*}}$ and the groundtruth ${y_i}$, where ${\widehat y_{is}}$ and ${\widehat y_{it}}$ are predictions of the student or teacher network, respectively.

\section{Experiments}
\subsection{Datasets and Implementation Details}
\noindent {\bf Datasets.} (1) UCF101 \cite{ucf101}. UCF101 dataset is collected from YouTube websites in unconstrained environments. It consists of 13320 videos with 101 action classes, including 23 hours of videos.
(2) HMDB51 \cite{hmdb51}. HMDB51 dataset contains $\approx$7000 videos collected from various sources, from movies to YouTube websites. It consists of 51 action classes, and each class has at least 101 videos.

(3) Something-Something V2 (Sth-v2)\cite{sthsth2020}. The Sth-v2 dataset is a large-scale crowd-sourced dataset containing rich human-object interactions. The dataset contains 174 categories and 220847 videos, among which 168913 videos are for the training set, 24777 videos are for the validation set, and 27157 videos are for the testing set. Since the testing set has no label, we use the validation set for reported results. Each video contains 2$\sim$6 seconds, and their frame rates are 12 fps/s.

{\bf Implementation Details.} We implement our models using PyTorch \cite{paszke2017automatic}, and we use SGD \cite{sgd} with a momentum rate of 0.9 to optimize our models on one GTX 3080Ti GPU. The learning rate is initialized to 0.00001 and is reduced by a multiplier of 0.95 at the $100$-th, $150$-th, $250$-th, and $350$-th epochs, respectively. The batch size is set to 16 on UCF101 and HMDB51 and 256 on Sth-v2. The dropout ratio is set to 0.5. The channels are set to 512 at the GC and DGC layers.
According to the official settings, the window size $P$ and the overlap of the sliding windows $O$ are set to 5 and 4 on the BN-Inception network, respectively. Similarly, they are set to 16 and 15 on the R(2+1)D and 3D ResNet networks, respectively. The height $H_i$ and width $W_i$ of the stacked images are set to 224 on BN-Inception networks and 112 on R(2+1)D and 3D ResNet networks, respectively.
Consistent with the baselines, we use the same training or testing splits on UCF101, HMDB51, and Sth-v2 for evaluations. {For a fair comparison, we strictly reproduce the results of PTS in section \ref{sec:4.3}.} The area under the curve (AUC) is used as our metric by average accuracies over different progress levels. {Consistent with the baselines, the number of progress levels $N$ is set to 10 on all datasets.} The number of stacked DGC layers (i.e., {$Q$}) is set to 1 on UCF101, HMDB51 and Sth-v2 datasets. We empirically set the maximum number of epochs for each dataset based on our empirical observations, i.e., 800 epochs on UCF101, 400 epochs on HMDB51, and 1000 epochs on Sth-v2, and then we choose the best model in terms of AUC.

\subsection{Baselines}
\noindent To show the effectiveness of our method, we choose several related methods as our baselines, including traditional methods, i.e., IBOW \cite{ryoo2011human}, DBOW \cite{ryoo2011human}, { MSSC \cite{cao2013recognize}}, {GLMP \cite{lai2017global}}, MTSSVM \cite{xu2019prediction}, and deep learning methods, i.e., DeepSCN \cite{kong2017deep}, Mem-LSTM \cite{kong2018action}, MSRNN \cite{hu2018early}, PTS \cite{wang2019progressive} AKT \cite{cai2019action}, AAPNet \cite{kong2020adversarial}, AJSM \cite{fernando2021anticipating}, AFR \cite{wu2021anticipating}, {IGGNN \cite{wu2021spatial}, ERA \cite{chen2022ambiguousness}, and TemPr \cite{stergiou2023wisdom}}. Moreover, AKT \cite{cai2019action}, PTS \cite{wang2019progressive}, {IGGNN \cite{wu2021spatial}, ERA \cite{chen2022ambiguousness}, and TemPr \cite{stergiou2023wisdom}} achieve current state-of-the-art performance.

\subsection{Comparison with baselines} \label{sec:4.3}
\begin{table*}
  \scriptsize
  \begin{center}
  \caption{Quantitative results on UCF101. The results for DBOW \cite{ryoo2011human}, IBOW \cite{ryoo2011human}, MTSSVM \cite{kong2014discriminative}, DeepSCN \cite{kong2017deep}, Mem-LSTM \cite{kong2018action}, MSRNN \cite{hu2018early} and PTS \cite{wang2019progressive} are reported in \cite{wang2019progressive}. The corresponding results of AKT \cite{cai2019action} are also reported in their paper. We reproduce the results of PTS using our features and fusion strategies, naming PTS$^i$ ($i$=1,2,3). ``*(*)'' reports the results with different features.
  The bold denotes the best results, and the underline denotes the second-best results. ``-'' denotes unavailable results. ``A\&B'' denotes using both A and B datasets, features, or modalities.}
  \centering
  \begin{tabular}{cccccccc}  
    \hline
    Method&Features&Pre-trained data&Modality&0.1&0.2&0.3&AUC\\
    \hline
    DBOW (ICCV2011) \cite{ryoo2011human} & STIP & No & RGB&36.29&51.57&52.71&51.37\\
    IBOW (ICCV2011) \cite{ryoo2011human} & STIP&No &RGB&36.29&65.69&71.69&70.01\\
    MTSSVM (ECCV2014)\cite{kong2014discriminative} & STIP\&DT& No& RGB & 40.05&72.83&80.02&77.41\\
    DeepSCN (CVPR2017) \cite{kong2017deep} &  C3D & Sports-1M  & RGB&45.02&77.64&82.95& 81.31 \\
    Mem-LSTM (AAAI2018) \cite{kong2018action} & ResNet-18 & ImageNet\&UCF101 & RGB\&flow& 51.02&80.97&85.73&84.10\\
    MSRNN (TPAMI2018) \cite{hu2018early} &  VGG & UCF101 & RGB\&flow &68.00&87.39&88.16& 87.25 \\
    AKT (AAAI2019) \cite{cai2019action} &  3D ResNeXt-101 &{ Kinetics-400} & RGB &80.00&84.7&86.90& 88.4\\
    \hline
    PTS (CVPR2019) \cite{wang2019progressive} & 3D ResNeXt-101 & { Kinetics-400} & RGB &83.32 &87.13&88.92&89.64\\
    {PTS$^{1}$ (CVPR2019) \cite{wang2019progressive}} & {BN-Inception} &{ Kinetics-400} & {RGB} &{82.24}&{85.42}& {86.77}&{87.23}\\
    {PTS$^{2}$ (CVPR2019) \cite{wang2019progressive}} & {BN-Inception} &{ Kinetics-400} & {flow} &{75.10}&{87.43}&{90.09}&{90.64}\\
     {PTS$^{3}$ (CVPR2019) \cite{wang2019progressive}} & {BN-Inception} &{ Kinetics-400} & {RGB\&flow} & {90.17}&{93.45} &{94.92} &{95.16}\\
     \hline
    AAPNet (TPAMI2020) \cite{kong2020adversarial} & BN-Inception & Unknown & RGB\&flow&59.84&80.44&86.78& 85.13\\
    AJSM (CVPR2021) \cite{fernando2021anticipating}& Resnet18(2D+1D)&Kinetics-400&RGB\&flow&{-} &91.70&{-}&{-}\\
    AFR (AAAI2021) \cite{wu2021anticipating} & BN-Inception& Kinetics-400\&UCF101&RGB\&flow&{82.36}&{85.57}&{88.97}&{-}\\
    {IGGNN (IJCV2021) \cite{wu2021spatial}} & {Faster R-CNN\&BN-Inception}  &{ImageNet\&Kinetics-400\&UCF101} & {RGB{\&}flow}&{80.26}& {--}&{89.86}&{--} \\
    {IGGNN (IJCV2021) \cite{wu2021spatial}} & {Faster R-CNN\&3D ResNext-101}  &{ImageNet\&Kinetics-400} & {RGB}& {80.24}&{--}& {84.55}&{--} \\
    {ERA (TCSVT2022) \cite{chen2022ambiguousness}} & {ResNeXt101} & {Unknown}& {RGB}& {89.10}&{--}&{92.40}&{--} \\
    {TemPr(1) (CVPR2023) \cite{stergiou2023wisdom}} & {3D ResNet-18} & {Kinetics-700\&UCF101}&{ RGB}& {84.30} &{90.20}& {90.40}&{--} \\
    {TemPr(2) (CVPR2023) \cite{stergiou2023wisdom}} & {3D ResNet-50} & {Kinetics-700\&UCF101}& {RGB}& {84.80} &{90.50} &{91.20}&{--} \\
    {TemPr(3) (CVPR2023) \cite{stergiou2023wisdom}} & {3D ResNeXt101} & {Kinetics-700\&UCF101}& {RGB}& {85.70}& {91.40}& {92.10}&{--} \\
    {TemPr(4) (CVPR2023) \cite{stergiou2023wisdom}} & {3D MoViNet-A4} & {Kinetics-700\&UCF101}& {RGB}& {88.60}& {93.50}& {94.90}&{--} \\
    \hline
    {Ours(1)} & {BN-Inception}&{ Kinetics-400} & {RGB}&{83.76}& {85.82}& {87.67}& {88.28} \\
    {Ours(2)} & {R(2+1)D-50}&{Kinetics-700} & {RGB}&{91.88}& {92.61}& {93.4}& {94.28} \\
    Ours(3) & 3D ResNet-18 & Kinetics-700 & RGB&84.49&85.47&86.31& 87.51 \\
    {Ours(4)} & {3D ResNet-34} &{Kinetics-700} & {RGB}&{87.32}& {88.38}& {89.71}& {90.81} \\
    {Ours(5)} & {3D ResNet-50} &{ Kinetics-700} & {RGB}&{87.56}&{87.59}&{89.38}& {90.44} \\
    {Ours(6)} & BN-Inception&{ Kinetics-400} & flow&73.06 &86.77 &90.36& 91.20 \\
    { Ours(7)} &{R(2+1)D-50\&3D ResNet-34}&{{Kinetics-700}}&{RGB\&RGB}&{\underline{92.42}}&{93.13}&{93.64}&{94.55} \\
    { Ours(8)} & {BN-Inception}&{ Kinetics-400} & {RGB{\&}flow}&{90.03}&{\underline{93.64}}&{\underline{95.08}}&{\underline{95.25}} \\
    { Ours(9)} &{BN-Inception\&R(2+1)D-50}&{Kinetics-400\&Kinetics-700}&{RGB\&flow}&{\bf 93.73}&{\bf 95.14}&{\bf 96.20}&{\bf 96.62} \\
    \hline
  \end{tabular}
  \label{tab:sota1}
\end{center}
\end{table*}

\begin{table*}
  \begin{center}
  \footnotesize
  \caption{Quantitative results on HMDB51. {The corresponding results of \cite{cao2013recognize,kong2015max,lai2017global} are estimated from the figs in \cite{cai2019action}. We reproduce the results of PTS using our features and fusion strategies, naming PTS$^i$ ($i$=1,2,3). PTS$^2$ uses a learning rate of 0.01. ``Ours(*)'' reports our results with different features.}
  The bold denotes the best results. ``A\&B'' denotes using both A and B datasets, features, or modalities.}
   \begin{tabular}{cccccccc}
    \hline
    Method&Features&Pre-trained data&Modality&0.1&0.2&0.3&AUC\\
    \hline
   { MSSC (CVPR2013) \cite{cao2013recognize}} & { Cuboids descriptors }& {No}& {RGB} &{12.00}& {18.00}&{24.00}&{30.55} \\
    {MTSSVM (TPAMI2015) \cite{kong2015max}} & {STIP{\&}DT}& {No} & {RGB} &{14.00} &{18.00} & {26.00} &{29.70} \\
   { GLMP (TIP2018) \cite{lai2017global}} & {STIP{\&}IDT} & {RGB}& {No} & {38.50} & {44.00} & {49.00} & {51.60} \\
    AKT (AAAI2019) \cite{cai2019action} &  3D ResNeXt-101 &{ Kinetics-400} & RGB &43.50&48.40&51.20&55.50\\
    \midrule
   {PTS$^{1}$(CVPR2019) \cite{wang2019progressive}} & BN-Inception &{ Kinetics-400} & RGB  & 46.87&50.47&53.24&55.64
 \\
    {PTS$^{2}$(CVPR2019) \cite{wang2019progressive}} & BN-Inception &{Kinetics-400} & flow  & 41.60&50.10&55.15&60.53 \\
    {PTS$^{3}$(CVPR2019) \cite{wang2019progressive}} & BN-Inception &{Kinetics-400} & RGB{\&}flow  & 53.52&58.47&61.69&65.89 \\
    \hline
    {IGGNN (IJCV2021) \cite{wu2021spatial}} & Faster R-CNN\&3D ResNext-101  &ImageNet\&Kinetics-400 & RGB& 45.10&{--}& 52.35&{--} \\
    \hline
    Ours(1) & BN-Inception&{ Kinetics-400} & RGB&46.44& 50.90 &54.89& 57.32\\
    Ours(2) & BN-Inception&{ Kinetics-400} & flow&41.18& 51.39& 54.93& 63.85\\
    Ours(3) & BN-Inception&{ Kinetics-400} & RGB{\&}flow&{\bf 53.56}&{\bf 58.92}&{\bf 62.58} &{\bf 68.55}\\
    \hline
  \end{tabular}
  \label{tab:sota2}
\end{center}
\end{table*}

\subsubsection{Prediction results on UCF101}
\noindent As shown in Table \ref{tab:sota1}, our method achieves state-of-art performance in terms of AUC. Moreover, our results with single-modality features (i.e., Ours(1)$\sim$Ours(6)) have exceeded most baselines, showing the effectiveness of our method. The multi-modal results that combine the results of different features achieve the best performance. These results confirm two things: firstly, RGB and flow modalities provide complementary information for accurate early predictions(e.g., Ours(1)/Ours(6) $vs.$ Ours(8), Ours(2)/Ours(6) $vs.$ Ours(9)); secondly, a two-stream framework with different structures of branches can effectively promote early predictions even if each branch uses the same modality data as the input (e.g., Ours(2)/Ours(4) $vs.$ Ours(7)).

Compared with the traditional methods with handcrafted features, such as \cite{ryoo2011human, kong2014discriminative}, we obtain a significant improvement, demonstrating the effectiveness of our method powerfully.
Compared with some typical methods with deep features \cite{kong2017deep,kong2018action,hu2018early}, we also achieve an appreciable performance by a margin of at least 15.31\% (96.62\% $vs$. 81.31\%). Moreover, Mem-LSTM \cite{kong2018action} and MSRNN \cite{hu2018early} pre-train their models on UCF101, which enables their models to obtain the domain action knowledge on UCF101. By contrast, we extract features of partial or full videos by the pre-trained model on Kinetics-400 without fine-tuning on UCF101. In this way, we can not obtain domain-specific knowledge. Nevertheless, we still achieve improved accuracy by an appreciable margin, showing the effectiveness of our method again.

Compared with other recent methods \cite{cai2019action,wang2019progressive,kong2020adversarial,fernando2021anticipating,wu2021anticipating,wu2021spatial,chen2022ambiguousness,stergiou2023wisdom}, we also achieve superior performance in terms of AUC. AFR \cite{wu2021anticipating} and IGGNN \cite{wu2021spatial} model spatio-temporal relationships among objects for early predictions, ignoring the scene features and the semantic knowledge among different partial videos. PTS \cite{wang2019progressive} distills progressive action knowledge between adjacent partial videos but ignores the correlations of partial videos between lower and higher progress levels. Similarly, TemPr \cite{stergiou2023wisdom} models fine-to-coarse action knowledge based on transformer but ignores distilling action knowledge from full videos.
AKT \cite{cai2019action}, AJSM \cite{fernando2021anticipating}, and AAPNet \cite{kong2020adversarial} transfer action knowledge from full to partial videos by learning the feature embedding and discriminative action classifiers. However, they ignore the correlations among partial videos with different progress levels, resulting in unsatisfactory performance.

\begin{figure*}[!t]
\includegraphics[width=0.99\linewidth,trim =30mm 68mm 30mm 5mm, clip=true]{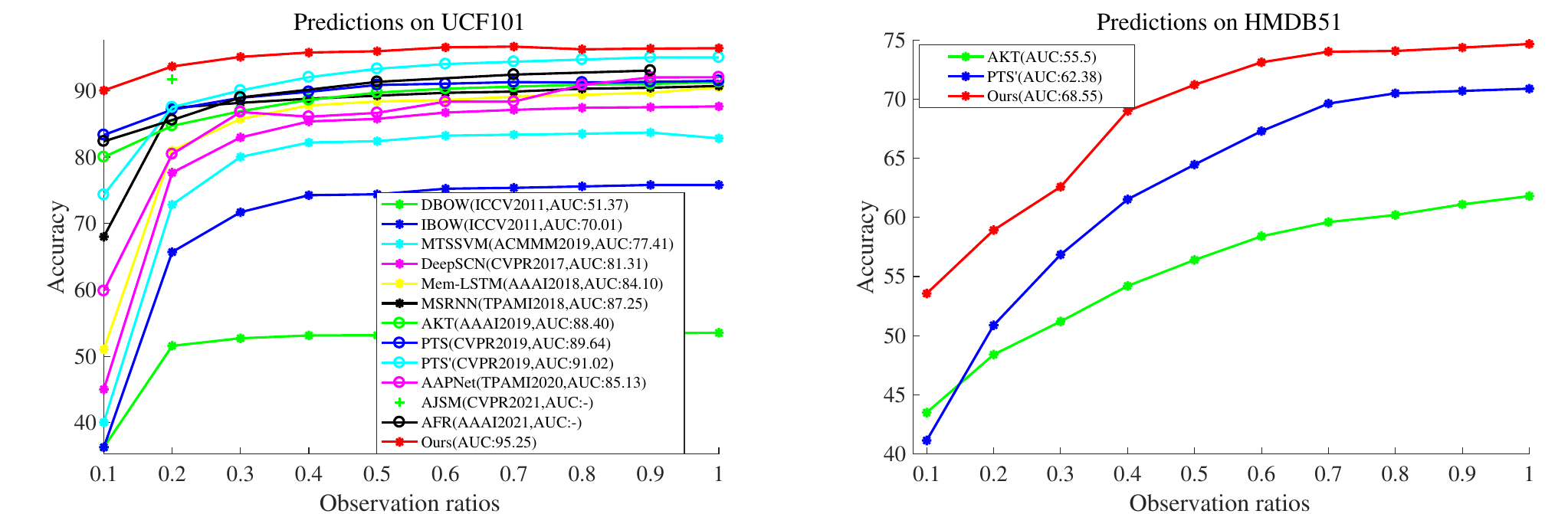}
\caption{Prediction results with different observation ratios on UCF101 and HMDB51 datasets. $(*)$ denotes the performance of the corresponding method in terms of AUC.}
\label{fig:sotawise}
\end{figure*}

We often ignore the phenomenon that any partial or full video usually contains other partial videos with lower progress levels than itself. Based on the observations, we model rich ASCK among arbitrary partial or full videos via graph-based knowledge reasoning and knowledge distillation. For the graph-based knowledge reasoning, we distill ASCK among partial videos from lower to higher progress levels and higher to lower progress levels by the teacher network; we distill ASCK of partial videos from lower to higher progress levels by the student network since future information is not available at the test phase. Under the knowledge distillation framework, we further transfer action knowledge from the teacher to the student networks, which enriches the ASCK of partial videos for accurate predictions. The above reasons possibly enable our model to achieve better predictions. We report detailed results with different observation ratios in Fig. \ref{fig:sotawise}. Our method outperforms the baselines at most progress levels, especially at very low observation ratios, showing our effectiveness again.

\begin{table*}
  \begin{center}
  \footnotesize
  \caption{Quantitative results on Sth-v2. For a fair comparison, we also reproduce the results of PTS \cite{wang2019progressive} using our extracted features, and we name it PTS'(*). ``Ours-RGB'' and ``Ours-flow'' report the results of ``RACK-RGB'' and ``RACK-flow'' subnetworks in Fig. \ref{fig:overallframework}, respectively.
  The bold denotes the best results. ``A\&B'' denotes using both A and B modalities.}
  \begin{tabular}{cccccccc}  
    \hline
    Method&Features &Pre-trained data&Modality&0.2&0.4&0.8&AUC\\
    \hline
    PTS'(1)(CVPR2019)\cite{wang2019progressive}& BN-Inception&{ Kinetics-400} & RGB&11.99& 14.00& 17.86& 15.25 \\
    PTS'(2)(CVPR2019)\cite{wang2019progressive} & BN-Inception&{ Kinetics-400} & flow&12.89& 20.31& 30.80& 23.36\\
    PTS'(3)(CVPR2019)\cite{wang2019progressive}& BN-Inception&{ Kinetics-400} & RGB\&flow&{\bf 16.92}& 23.08& 32.43& 25.91\\
    \hline
    Ours-RGB & BN-Inception&{ Kinetics-400} & RGB&11.90& 15.04 &23.02& 18.24\\
    Ours-flow & BN-Inception&{Kinetics-400} & flow&12.83& 20.85& 32.98& 24.56\\
    Ours & BN-Inception&{ Kinetics-400} & RGB\&flow&{16.21}&{\bf 23.67}& {\bf 35.51}&{\bf 27.63}\\
    \hline
  \end{tabular}
  \label{tab:sota3}
\end{center}
\end{table*}

\begin{figure*}[!t]  
\begin{center}
\includegraphics[width=0.99\linewidth,trim =35mm 10mm 30mm 15mm, clip=true]{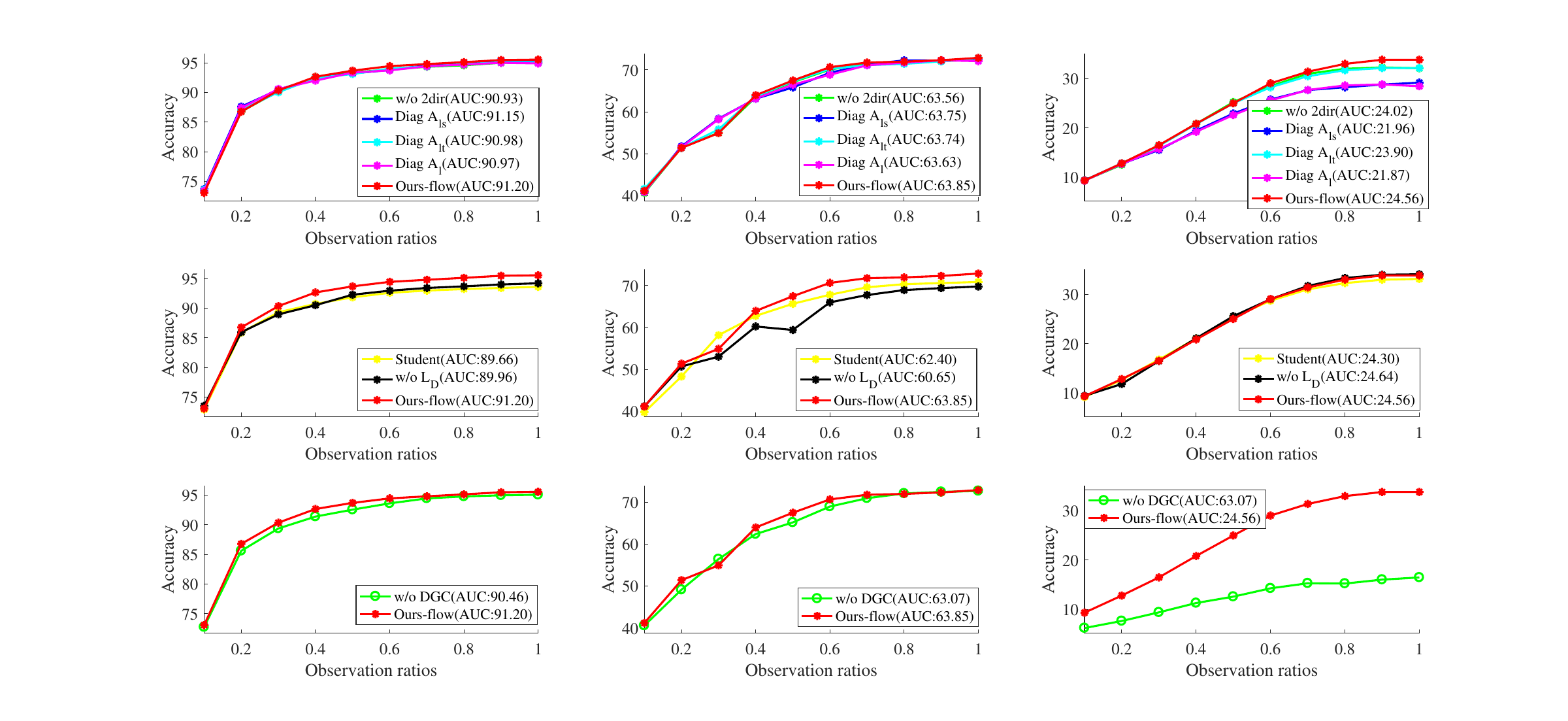}
\end{center}
\caption{Detailed results of ablative studies. $(*)$ denotes the performance of the corresponding method in terms of AUC. The left one denotes the results on UCF101, the middle one denotes the results on HMDB51, and the right one denotes the results on Sth-v2.}
\label{fig:ab}
\end{figure*}

\begin{table*}
  \begin{center}
  \footnotesize
  \caption{Parameter analysis. The bold denotes the best results. ``{+*}'' denotes the improvement in terms of AUC.}
  \begin{tabular}{cccccc}
    \hline
    \multirow{2}{*}{Method} & \multirow{2}{*}{GPU Memory}& \multirow{2}{*}{{Parameters}} &\multicolumn{2}{c}{AUC}\\
    \cline{4-6}&&&UCF101&HMDB51&{Sth-v2}\\
    \hline
    PTS'{(CVPR2019)\cite{wang2019progressive}} &3303M&5.8M&91.02&62.38&23.36\\
    {RACK-flow(Ours)} & 861M&2.2M& 91.20(+0.18)&63.85(+1.47)&24.56(+1.2) \\
     \hline
  \end{tabular}
  \label{tab:sota4}
\end{center}
\end{table*}

\subsubsection{Prediction results on HMDB51}
\noindent Compared with the UCF101 dataset, the HMDB51 dataset is more challenging due to more diverse sources and larger intra-class variances, making it more difficult to predict human action very early. As shown in Table \ref{tab:sota2}, the performance of our method significantly outperforms that of traditional methods \cite{cao2013recognize,kong2015max,lai2017global}.
Compared with AKT \cite{cai2019action} and PTS$^2$ \cite{wang2019progressive}, we achieve superior performance in terms of AUC by a margin of 13.05\% and 2.66\%, respectively.
Although \cite{cai2019action} and \cite{wang2019progressive} try to distill action knowledge from full videos, they ignore ASCK among partial videos with different progress levels, especially the semantic consistencies between higher and lower progress levels. {Compared with the recent method \cite{wu2021spatial}, we also achieve improved accuracy. \cite{wu2021spatial} focuses on modeling spatio-temporal relationships among objects and persons.}
In contrast, we model rich ASCK among partial videos with richer progress levels via graph-based knowledge reasoning and knowledge distillation. The mentioned reason may contribute to our superior performance on the more challenging dataset, strongly showing the importance of explicitly modeling richer ASCK among partial videos for complex activities.

Fig. \ref{fig:sotawise} shows the detailed results with different observation ratios on HMDB51. We achieve the highest accuracy by an appreciable margin at all observation ratios. Specifically, compared with AKT \cite{cai2019action} and PTS' \cite{wang2019progressive}, we obtain an improved accuracy at the progress level of 1 by 10.06\% and 12.42\%, respectively, and our performance is consistently improved at the following progress levels.

\subsubsection{Prediction results on Sth-v2}
\noindent Unlike UCF101 and HMDB51 datasets that contain rich scene information, the Sth-v2 dataset contains richer motion information, and it is hard to predict human actions using the scene information. Therefore, predicting human action very early is particularly challenging. It is more important to capture {ASCK} for predictions on such a dataset. Since the existing related work on the Sth-v2 dataset has not been evaluated, and their source code is not open source, we have utilized our extracted RGB and flow features along with our fusion strategies to replicate the results of PTS \cite{wang2019progressive} as a baseline. Detailed results are shown in Table \ref{tab:sota3}. Compared with PTS', our method consistently outperforms it using the same features in terms of AUC (15.25\% $vs$. 18.24\%, 23.36\% $vs$. 24.56\%, 25.91\% $vs$. 27.63\%). For different observation ratios, we also achieve better results in most cases. For the observation ratio of 0.2, our accuracy is slightly decreased. The possible reason is the limited useful information and low predictability of actions caused by short videos at a very early. At the later observation ratios, we achieve improved performance significantly, showing the effectiveness of our method.

\subsection{Ablation analysis}
\noindent In this section, we conduct extensive experiments to show the effectiveness of our key contributions. Firstly, we first verify the effect of different features. Then, we show the effectiveness of rich ASCK via graph-based knowledge reasoning and knowledge distillation. Next, we evaluate the rationality of the DGC structure. Finally, we briefly evaluate our knowledge distillation-based training efficiency and hyperparameters.

\subsubsection{{Evaluation the effect of different features}}
\noindent To evaluate the effect of different features, we extract video features using different backbones with various depths\footnote{{Due to the time-consuming of calculating optical flow, here we mainly use RGB for experiments}}, including R(2+1)D \cite{tran2018closer}, 3D ResNet \cite{hara2018can} and BN-Inception \cite{tsn}.

Detailed results are shown in Table \ref{tab:abfev}. Regarding single modality, R(2+1)D-50 achieves the best performance with an AUC of 94.28\%. Overall, the improved performance increases with depths of the same backbone, and the R(2+1)D achieves the best performance. However, when the network reaches a certain depth, the improved performance is limited, such as ``3D ResNet''. In terms of two-stream fusion, two-stream fusion can significantly improve the predictions by using different architectures or different modalities, among which RGB and flow fusion obtain better performance, such as ``BN-Inception\&R(2+1)D-50'' with an AUC of 96.62\%.\footnote{Further results can be found at our official code repository.} The results show that flow modality can provide complementary information for accurate predictions.
Moreover, the feature dimension $D$ does not have an immediate impact on model performance. The specific network structure and its ability to extract robust features play a crucial role in determining the performance of the model. For example, the features with the same dimensions have various performances, which may be affected by the depth and network structure of the backbone, e.g., 3D ResNet-18 $vs.$ ResNet-34 (87.51\% $vs.$ 90.81\%), 3D ResNet-18 $vs.$ R(2+1)D-18 (87.51\% $vs.$ 93.05\%). The features with increased dimensions may have limited improved performance, e.g., R(2+1)D-34 $vs.$ R(2+1)D-50 (93.35\% $vs.$ 94.28\%), 3D ResNet-34 vs 3D ResNet-50 (90.81\% $vs.$ 90.44\%).

\begin{table}
\begin{center}
  \scriptsize
  \caption{{Quantitative results of different features on UCF101.``$A$-$B$'' in the first column denotes the backbone $A$ with a depth of $B$. The bold denotes the best results, and the underline denotes the second-best results.} ``A\&B'' denotes using both A and B features or modalities.}

  \begin{tabular}{ccccc}  
    \hline
    Features& {Dimension} &Modality&0.1&AUC\\
    \hline
    R(2+1)D-18& {512}  &RGB& 90.33   & 93.05 \\
    R(2+1)D-34 & {512} &RGB& 90.93& 93.35 \\
    R(2+1)D-50 &{2048} &RGB& \underline{91.88}& 94.28 \\
    3D ResNet-18& {512}  & RGB& 84.49&  87.51 \\
    3D ResNet-34 &{512} & RGB&87.32 &  90.81 \\
    3D ResNet-50 & {2048} & RGB& 87.56&   90.44 \\
    3D ResNet-101 & {2048}& RGB&87.81 &    90.82 \\
    3D ResNet-152 & {2048}& RGB& 87.86&   90.77 \\
    3D ResNet-200 & {2048} & RGB&88.08 &   91.18 \\
    BN-Inception & {1024} & RGB&83.76 &  88.28  \\
    BN-Inception  & {1024}& flow&86.77 & 91.20  \\
    \hline
    BN-Inception  & {1024}& RGB{\&}flow& 90.03& 95.25\\
    R(2+1)D-50\&3D ResNet-200& {--} &  RGB{\&}RGB & 91.96&94.12\\
    BN-Inception\&R(2+1)D-50 & {--}& RGB{\&}flow &{\bf 93.73}&{\bf 96.62}\\
    BN-Inception\&3D ResNet-200& {--} & RGB{\&}flow &91.42&\underline{96.21}\\
    \hline
  \end{tabular}
  \label{tab:abfev}
\end{center}
\end{table}

\subsubsection{Evaluation of modeling rich ASCK}
\noindent Our rich ASCK modeling benefits from two-fold: graph-based knowledge reasoning and knowledge distillation.

{\bf Effectiveness of graph-based knowledge reasoning.} We conduct four group experiments to show the effectiveness of graph-based knowledge reasoning. (1) Change the bi-directional graph of the teacher network into a single directional graph, denoting by ``w/o 2dir''. In this way, the teacher network models action knowledge from lower to higher progress levels. (2) Setting the adjacent matrix of the teacher network to a diagonal matrix. In this way, the teacher network ignores modeling ASCK of different progress levels, denoting by ``Diag ${{\bf{\emph A}}_{lt}}$''. (3) Setting the adjacent matrix of the student network to a diagonal matrix. In this way, the student network ignores modeling ASCK of different progress levels, denoting by ``Diag ${{\bf{\emph A}}_{ls}}$''. (4) Setting the adjacent matrix of the teacher and student networks to a diagonal matrix. In this way, the teacher and student networks ignore modeling ASCK of different progress levels, denoting by ``Diag ${{\bf{\emph A}}_{l}}$''.

As shown in Table \ref{tab:ab}, the top ones show the effectiveness of distilling ASCK via graph-based knowledge reasoning. Compared with ``Ours-flow'', the AUC of ``w/o 2dir'' decreases by 0.27\%, 0.29\% and 0.54\% on UCF101, HMDB51, and Sth-v2, respectively.
The single-directional graph by the teacher network still enables the model to distill ASCK from lower to higher progress levels. Therefore, the teacher network with the single-directional graph also greatly helps to improve the predictions. When we do not consider the ASCK of the teacher or student network by setting the adjacent matrixes to the diagonal matrixes (i.e., ``Diag ${{\bf{\emph A}}_{ls}}$'', ``Diag ${{\bf{\emph A}}_{lt}}$'', and ``Diag ${{\bf{\emph A}}_{l}}$''), the $GC$ layer in equation (1) becomes a fully connected layer, which means that the model will ignore modeling ASCK between partial videos. As a result,
the performance consistently decreases, especially on Sth-v2. The possible reasons for this are two-fold: (1) UCF101 and HMDB are scene-related datasets. Partial videos that contain scene context and short-term dynamics may be sufficient for predictions. (2) Sth-v2 is a motion-related dataset. Partial videos contain limited information on action semantics. Compared to the task on UCF101 and HMDB51, early predictions on Sth-v2 are quite challenging. It is crucial to model temporal dynamics by graph knowledge reasoning, especially long-term dynamics, for predictions. This may explain why the improved accuracy on UCF101 and HMDB51 is limited while significant improvement is observed on Sth-v2.

{\bf Effectiveness of knowledge distillation.} We conduct two group experiments to show the effectiveness of modeling rich ASCK via knowledge distillation. (1) Removing the teacher network, denoting by ``Student''. In this way, our model ignores the ASCK from the full videos. (2) Removing the feature-level distillation loss $L_{D}$ in equation \ref{eq7}, denoting by ``w/o $L_{D}$''. In this way, our model will ignore learning feature-level ASCK from the teacher to the student network.

 Detailed results are shown in Table \ref{tab:ab}. Without considering the ASCK from the teacher network (``Student''), the AUC decreases by 1.54\%, 1.45\%, and 0.26\% on UCF101, HMDB51, and Sth-v2, respectively. Compared with ``Ours-flow'', the AUC of ``w/o $L_D$'' also decreases by 1.24\% and 3.2\% on UCF101 and HMDB51, respectively. These results prove the effectiveness of rich action knowledge modeling via knowledge distillation, especially on scene-related datasets (i.e., UCF101 and HMDB51).
The slightly improved 0.08\% without $L_D$ on the Sth-v2 dataset can largely be attributed to the unique characteristics of this dataset. The Sth-v2 dataset is a motion-related dataset that contains abundant motion information, making it more crucial to extract motion dynamics about actions for early recognition. As discussed above, $L_D$ attempts to constrain the model to extract scene-related features from full videos. However, scene-related features may not contribute to the early predictions on the Sth-v2. Instead, they may introduce irrelevant noise, leading to a slightly declined performance.

\begin{table}
  \small
  \begin{center}
  \caption{Results of ablative studies. The bold marks the best results, ``(-*)'' shows the decreased accuracy compared with our results, and the AUC denotes the average prediction overall ten progress levels.}

  \begin{tabular}{cccc}  
    \hline
    \multirow{2}{*}{Model}&\multicolumn{3}{c}{AUC} \\
    \cline{2-4}& UCF101& HMDB51&Sth-v2\\
    \hline
    w/o 2dir &90.93(-0.27)& 63.56(-0.29)& 24.02(-0.54)\\
    Diag ${{\bf{\emph A}}_{ls}}$ & 91.15(-0.05)& 63.75(-0.10)& 21.96(-2.60)\\
    Diag ${{\bf{\emph A}}_{lt}}$ & 90.98(-0.22) &63.74(-0.11) &23.90(-0.66)\\
    Diag ${{\bf{\emph A}}_l}$ & 90.97(-0.23)& 63.63(-0.22)& 21.87(-2.69)\\
    \hline
    Student &  89.66(-1.54)& 62.40(-1.45) &24.30(-0.26)\\
    w/o $L_D$ & 89.96(-1.24)& 60.65(-3.2)& {\bf 24.64(+0.08)}\\
    \hline
    w/o DC & 90.46(-0.74)& 63.07(-0.78)& 12.47(-12.09)\\
    Ours-flow &{\bf 91.20}& {\bf 63.85}& 24.56\\
    \hline
    \end{tabular}
  \label{tab:ab}
\end{center}
\end{table}

Fig. \ref{fig:ab} shows the detailed results with different observation ratios on UCF101, HMDB51, and Sth-v2. The improved results on various datasets show the effectiveness of our rich ASCK modeling via graph-based knowledge reasoning and knowledge distillation. Comparing their results, we make two interesting conclusions. (1) It is more effective to model rich ASCK via knowledge distillation for the predictions on the scene-related datasets (i.e., UCF101 and HMDB51).
Of course, the knowledge distillation framework may introduce additional noises, such as the background and semantic noises, for the dynamic backgrounds and complex activities, leading to incorrect predictions, such as an observation rate of 0.3 on HMDB51 (``Ours-flow'' $vs.$ ``Student'').
(2) while it is more effective to model rich ASCK via graph-based knowledge reasoning for the predictions on the motion-related dataset (i.e., Sth-v2).

\subsubsection{Evaluation of the rationality of DGC structure}
\noindent ``w/o DC'' denotes without densely residual connections in the DGC block, i.e., we remove the blue lines of DGC layers in Fig. \ref{fig:overallframework}. As shown in Table \ref{tab:ab}, the performance of ``w/o DC'' consistently decreases on three datasets, particularly by 12.09\% in terms of AUC on Sth-v2.
Without residual connections, the node features of partial video cannot be enhanced from low-level features, and their features may degrade with the increasing depth of the network.
Moreover, the residual connections potentially encourage models to learn motion features of human actions, which is critical for motion-related datasets such as Sth-v2. This has been well-established in human motion prediction, which is commonly used to forecast the velocities of human motion sequences\cite{hmprnn}. The above reasons may cause a significant decline in performance on the Sth-v2 dataset without residual connections.
The detailed results in Fig. \ref{fig:ab} also show similar improved results. Therefore, the results in Table \ref{tab:ab} and Fig. \ref{fig:ab} consistently show the rationality of the DGC structure.

\subsubsection{{Evaluation of knowledge distillation-based training efficiency}}
\noindent To evaluate the knowledge distillation(KD)-based training efficiency, we conduct experiments using PTS and RACK-flow models on UCF101. As shown in Table \ref{tab:abtrain}, our method has lower GFLOPs and parameters than PTS. The lower GFLOPs and parameters may benefit from our simple graph model that only contains three graph layers, showing the lower complexity of our method. However, we obtain higher training time per epoch. The probable cause could be the large matrix computations performed in the convolution layer, and the large matrix computations are time-consuming, leading to higher training time.

Compared with ``RACK-student (Ours)'', the GFLOPs, parameters, and training time of ''RACK-flow (Ours)'' are doubled due to the KD-based framework. However, our model is lightweight with parameters of 2.24M, and the increased GFLOPs and parameters can be ignored. Thus, the increased training time may be tolerant.

\begin{table}
  \footnotesize
  \caption{{Quantitative results of KD-based training efficiency on UCF101. ``PTS'' and ``RACK-flow (Ours)'' denote teacher and student networks, ``PTS-student'' and ``RACK-student (Ours)'' denotes only the student network. ``Time'' denotes the training time per epoch.}}
  \centering
  \begin{tabular}{ccccc}
    \hline
    Method & GFLOPs& Parameters& Time & AUC \\
    \hline
    PTS \cite{wang2019progressive}& 116.7M&	5.83M&	0.8s&	90.64 \\
    RACK-flow (Ours)&	2.24M	&2.19M	&4.6s	&91.20 \\
    PTS-student \cite{wang2019progressive}&	64.12M	&3.20M&	0.5s	&89.70  \\
    RACK-student (Ours)&	1.12M	&1.12M	&2.4s&	89.66  \\
   \hline
  \end{tabular}
  \label{tab:abtrain}
\end{table}

\subsubsection{Hyperparameter analysis}
This section demonstrates the effectiveness of hyperparameter selections, including the similarity measures for matrix $S$ in equation \ref{eq4}, the number of stacked DGC layers (i.e., $Q$), and weights of $L_D$ and $L_C$.

\begin{table}
  \small
  \caption{The evaluation of Hyperparameters. ``PCC'' denotes Pearson correlation coefficient, ``$L_2$'' denotes Euclidean distance, and ``CS'' denotes cosine similarity. The bold denotes the best results, and the underline denotes the second-best results. ``-'' denotes unavailable results.``Time'' denotes the training time per epoch.}
  \centering
  \begin{tabular}{ccc}
    \hline
    \multicolumn{3}{c}{The evaluation of different metrics on UCF101} \\
    Metrics &AUC& Time \\
    \hline
     $L_2$ &80.36& \underline{3664.38s}\\
     PCC &\underline{90.97}& 3738.50s\\
     CS(Ours-flow) &{\bf 91.20}& {\bf 35.16s}\\
     \hline
     \hline
     \multicolumn{3}{c}{The evaluation of different number of DGC layers (i.e., $Q$)} \\
     $Q$ & {AUC on UCF101}&{AUC on Sth-v2} \\
    \hline
    $1$(Ours-flow) &{\bf 91.20}&{\bf 24.56}\\
    $2$ &{\underline{90.82}}&{\underline{23.81}}\\
    $3$ &{89.95}&{--}\\
    $4$ &{90.09}&{--}\\
    $>4$ &{--}&{--}\\
    \hline
    \hline
    \multicolumn{3}{c}{The evaluation of weights for $L_D$ and $L_C$ losses} \\
     $L_C$:$L_D$ & {AUC on UCF101}&{AUC on Sth-v2} \\
     \hline
     1:0.1&{{\bf 91.26}}&{\underline{24.14}} \\
     1:0.01&{90.88}&{24.00} \\
     1:1(Ours-flow)&\underline{91.20}&{\bf 24.56} \\
     $w_1$:$w_2$&{\underline{91.20}}&{23.11} \\
   \hline
  \end{tabular}
  \label{tab:hyper}
\end{table}

{\bf The evaluation of different metrics for $S$ in equation \ref{eq4}.} Cosine similarity is a commonly used measure of similarity that can be applied to calculate the semantic correlation between different partial videos. Furthermore, cosine similarity is bounded within the range of [-1,1], which makes it easier to optimize.

To further validate the effectiveness of cosine similarity, we replace it with commonly used similarity metrics, including Euclidean distance ($L_2$) and Pearson correlation coefficient (PCC). As shown in Table \ref{tab:hyper}, CS outperforms $L_2$ in terms of AUC, while PCC performs similarly to CS. This demonstrates the effectiveness of using CS to measure the semantic correlations between different partial videos. Additionally, PCC has a boundary that makes it easier to optimize, while $L_2$ does not have an upper bound and is more challenging to train. In terms of training time per epoch, CS achieves the lowest time, while PCC and $L_2$ have higher times. This may be because CS can be accelerated through matrix operations, whereas PCC or $L_2$ calculates the semantics of any pairwise videos. Therefore, when dealing with a large number of videos, both PCC and $L_2$ can be time-consuming.
Therefore, we choose CS as our similarity measure for implementing matrix $S$ in equation \ref{eq4}.

{\bf The evaluation of different numbers of DGC layers.} We set $Q$ from 1 to 7 to demonstrate the effect of different stacked lengths of DGC layers. As shown in Table \ref{tab:hyper}, one DGC layer achieves the best performance on the three datasets. However, as we increase the depth of the network, the model may suffer from decreased performance or fail to converge. (1)
The results on Sth-v2 are not available when $Q>2$. In contrast to UCF101, which contains rich scene context, Sth-v2 contains rich motion information of human actions and exhibits more significant intra-class variability. Consequently, even with an increasing numbers of graph layers (e.g., $Q=3$ or $Q=4$), the dataset's intricate semantic ambiguity suffers from challenges for model optimization, particularly for partial videos with low observation ratios, potentially resulting in non-convergence of the model.
(2) When $Q>4$, our experiments on both UCF101 and Sth-v2 datasets yield non-convergent results, rendering them unavailable. One possible explanation is that our model has few nodes (i.e., 10), and perhaps a shallow GCN is sufficient to model the relationships between different nodes. A deep GCN may suffer from over-smoothing\cite{huang2020tackling,2020graph} or vanishing/exploding gradients\cite{li2019deepgcns}, potentially leading to training difficulties for model convergence. Finally, we set $Q$ to 1 on the three datasets.

{\bf The evaluation of weights for $L_D$ and $L_C$ losses.} To evaluate the impact of different loss weights, we set the weight ratios of $L_C$ and $L_D$ to be: 1:0.1, 1:0.01, 1:1, and $w_1$:$w_2$, respectively, where $w_1$ and $w_2$ are two learnable weights. As shown in Table \ref{tab:hyper}, for the UCF101 dataset, the weight ratio of 1:0.1 achieves the best performance, followed by the weight ratios of 1:1 and $w_1$:$w_2$; for the Sth-v2 dataset, the weight ratio of 1:1 achieves the best performance. Considering both performance and parameter saving, we choose a weight ratio of 1:1 by simply adding the $L_D$ and $L_C$ losses to optimize our model.

\subsection{Visualization}
\begin{figure}[!t]  
\begin{center}
\subfloat[Visualization using RGB features.]{\includegraphics[width=0.95\linewidth,trim =30mm 5mm 15mm 2mm, clip=true]{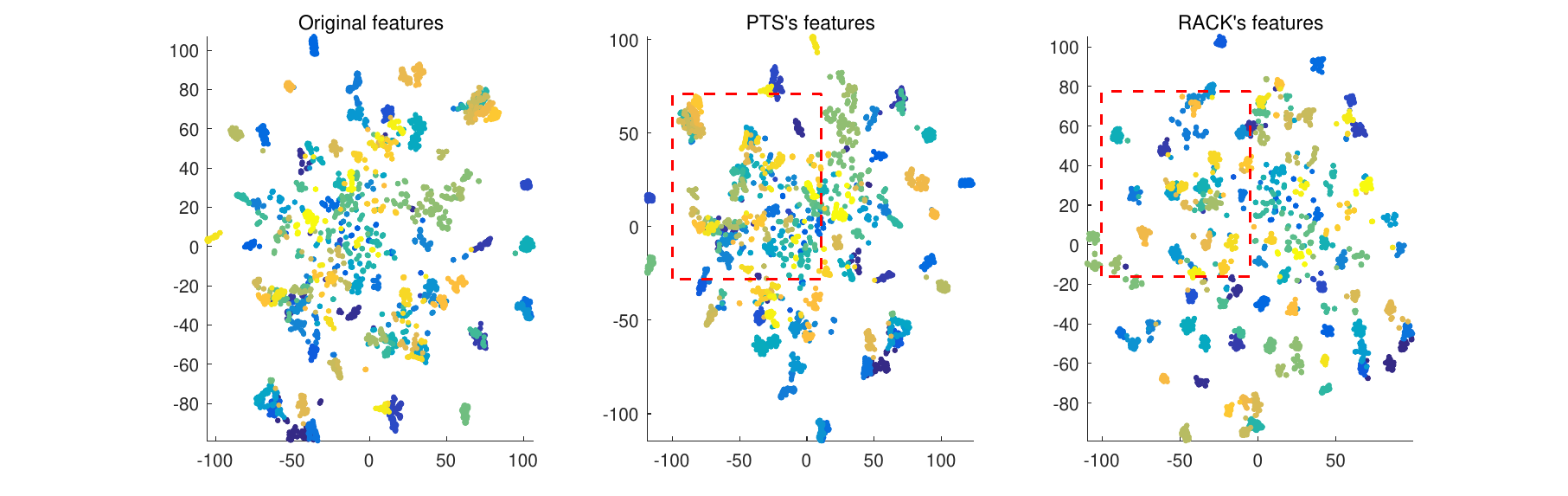}
\label{fig:visf1}
}
\\
\subfloat[Visualization using flow features.]{\includegraphics[width=0.95\linewidth,trim =30mm 5mm 15mm 2mm, clip=true]{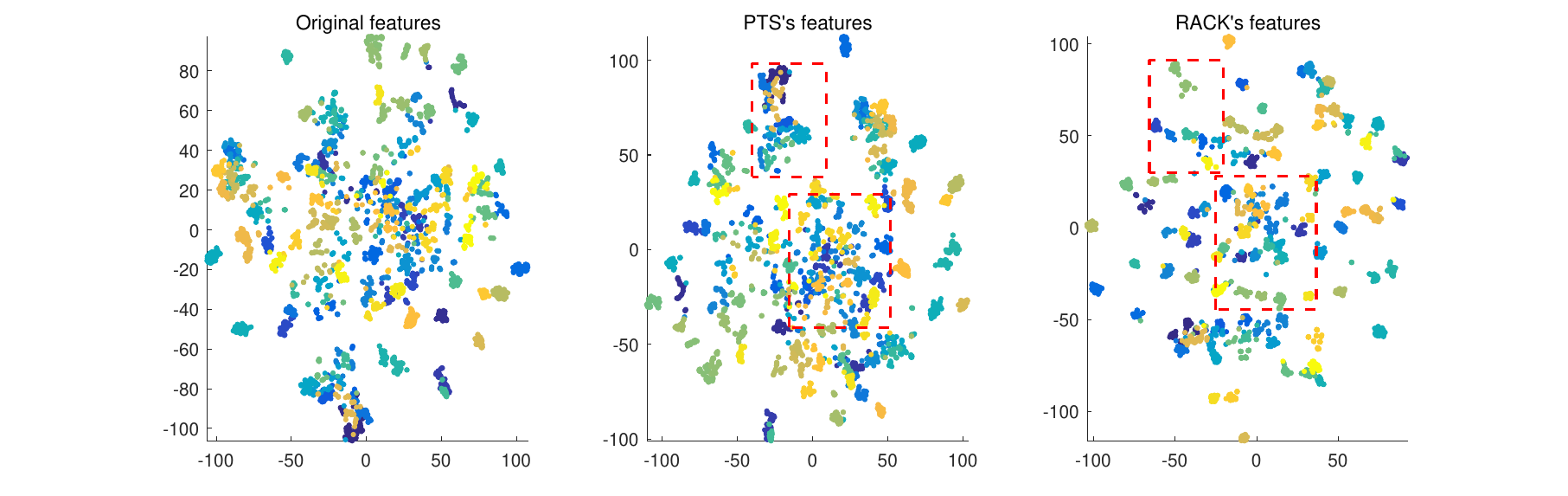}
\label{fig:visf2}
}
\end{center}
\caption{$t$-SNE\cite{van2008visualizing} visualizations at an observation ratio of 0.5 on UCF101.}
\label{fig:visf}
\end{figure}

\noindent To further show the effectiveness of our proposed RACK network, we visualize our learned features {and PTS' features} using the $t$-SNE technique \cite{van2008visualizing}. Taking an example of flow modality, we treat the extracted features of partial videos in section \ref{sec:3.3} as the original features, and we treat the features before the FC layer of the student network at the RACK-flow in Fig. \ref{fig:overallframework} as our learned features. {Similarly, we treat the features before the FC layer of the student network as PTS's features.
As denoted in Fig. \ref{fig:visf}, we respectively visualize the RGB and flow features of partial videos with an observation ratio of 0.5. The visualizations show that our learned features can be grouped better, indicating the effectiveness of rich ASCK by our proposed RACK network.}

\begin{figure}[!t]  
\begin{center}
\subfloat[Failed results on UCF101.]{\includegraphics[width=0.95\linewidth,trim =188mm 22mm 123mm 12mm, clip=true]{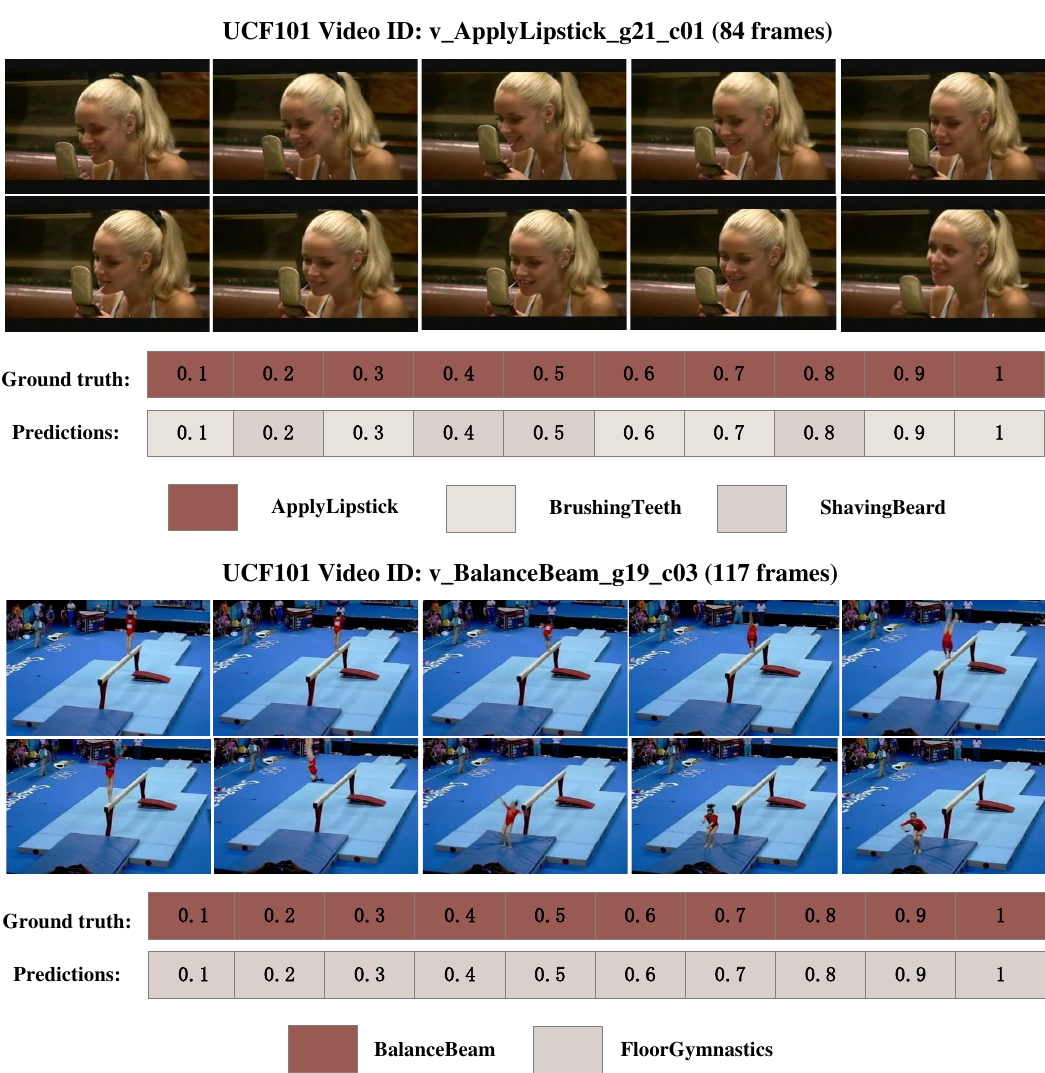}
\label{fig:vis1}
}
\\
\subfloat[Failed results on Sth-v2.]{\includegraphics[width=0.95\linewidth,trim =18mm 88mm 13mm 10mm, clip=true]{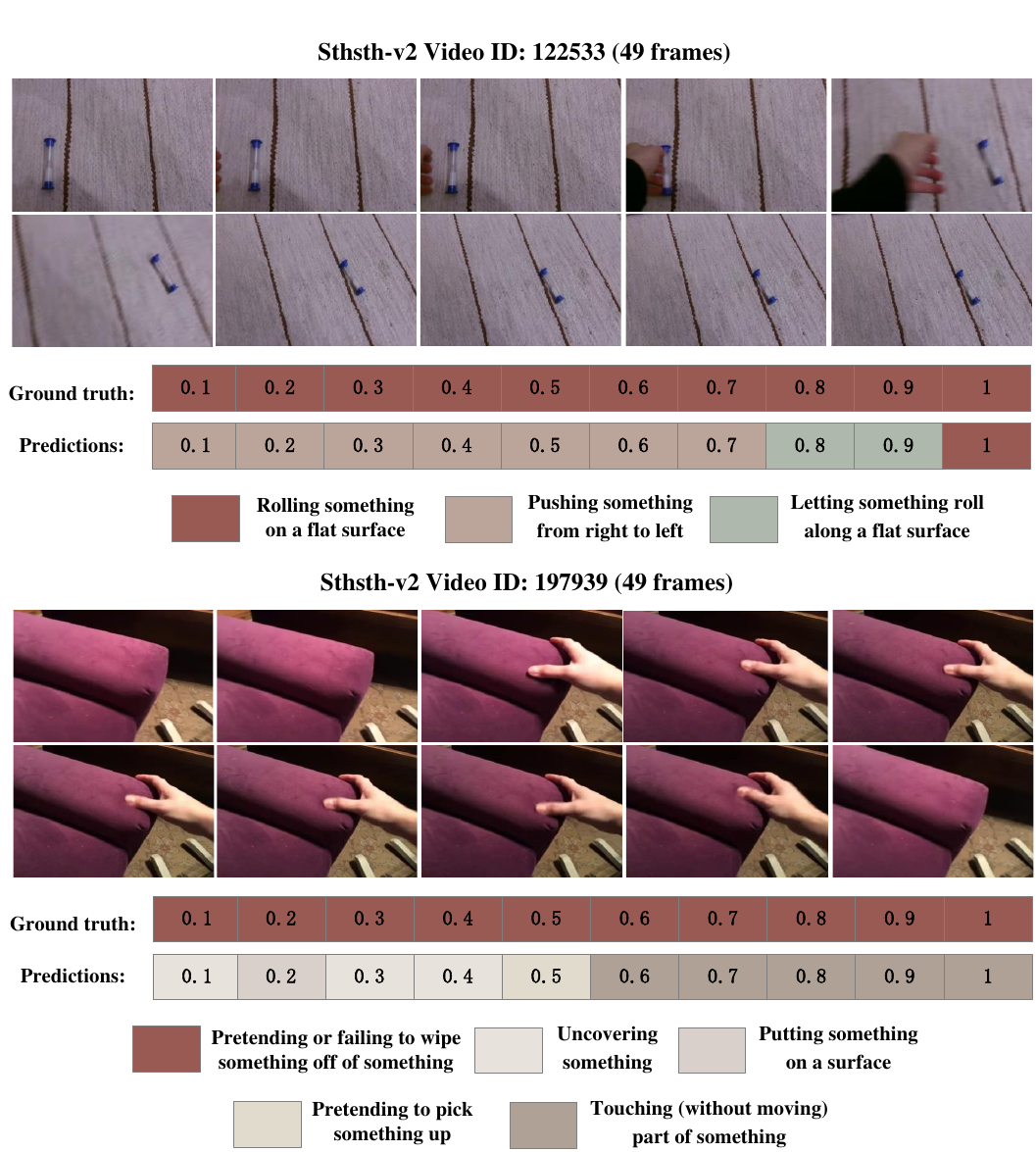}
\label{fig:vis2}
}
\end{center}
\caption{Visualization of failed predictions. The selected images are uniformly sampled from original videos. The boxes in different colors denote the predictions or Ground truth.}
\label{fig:vis}
\end{figure}

\subsection{Failed cases analysis}
\noindent To reveal the limitation of our proposed method, we visualize some failed predictions on UCF101 and Sth-v2 datasets, as shown in Fig. \ref{fig:vis}. The failed cases on UCF101 are shown in Fig. \ref{fig:vis}(a). (1) ``ApplyLipstick'' is easily predicted as ``BrushingTeeth'' and ``ShavingBeard''. Two main reasons may lead to the failed results. One is the similar spatio-temporal semantics of these activities. The other is short durations of full videos that may lead to extremely limited action information in partial videos. (2) ``BalanceBeam'' is incorrectly predicted as ``FloorGymnastics''. The possible reason is that the human body movements of ``BalanceBeam'' and ``FloorGymnastics'' are very similar. They need to be combined with a strong scene context before better distinguishing. However, we find that the scene context of ``BalanceBeam'' and ``FloorGymnastics'' are usually confused on the training set. These reasons may lead to incorrect predictions about the two activities.

As shown in Fig. \ref{fig:vis}(b), the failed cases on Sth-v2 are similar to those on UCF101, and they are usually misjudged as activities with similar spatio-temporal semantics and short durations. Another important reason is that the pre-trained Kinetics-400 dataset is a scene context-related dataset, while Sth-v2 is a motion dataset. This may lead to the domain gap between different datasets (scene dataset $vs.$ motion dataset) and various tasks (action recognition $vs.$ action prediction). Therefore, the extracted features may not sufficiently represent the actions of the Sth-v2 dataset and have poor predictive ability, especially for these highly similar activities.

In summary, although our proposed method has achieved a promising performance on multiple diverse datasets, there are some limitations on incorrect predictions on the activities with complex coupled backgrounds, extremely short durations, and highly similar action spatio-temporal semantics. In the future, we will explore a domain feature representation and a hierarchical prediction scheme for these similar activities.

\section{Conclusion}
\noindent In this paper, we propose a novel RACK network under a teacher-student framework, mining rich ASCK of partial videos with arbitrary progress levels utilizing graph-based knowledge reasoning and knowledge distillation. Specifically, we build a bi-directional and a single-directional fully connected graph for the teacher and student networks, respectively. The teacher network with the bi-directional graph considers ASCK of partial videos with arbitrary progress levels. The student network with the single-directional graph captures ASCK among partial videos from lower to higher progress levels. Also, it enriches their ASCK from the teacher network via distillation loss.
In contrast to prior works, we model rich ASCK among arbitrary partial videos and distill ASCK by graph-based knowledge reasoning and the teacher-student framework. Extensive experiments have shown the effectiveness of explicitly modeling rich ASCK for predictions.
Moreover, the proposed method can be easily extended to other high-level reasoning tasks, such as visual or dialog reasoning \cite{liang2022visual,zheng2019reasoning}, knowledge reasoning \cite{chen2020review}, and so on. However, the proposed method fails to predict similar activities with short durations and confusing backgrounds. We will explore a more profound investigation in the future.

\ifCLASSOPTIONcompsoc
  \section*{Acknowledgments}
\else
  \section*{Acknowledgment}
\fi

This work was supported partly by the National Natural Science Foundation of China (Grant No. 62173045, 61673192), partly supported by the Fundamental Research Funds for the Central Universities(Grant No. 2020XD-A04-2), and partly supported by BUPT Excellent Ph.D. Students Foundation (CX2021314).

\bibliographystyle{IEEEtran}
\bibliography{RACK_arXiv}

\begin{IEEEbiography}[{\includegraphics[width=1in,height=1.25in,clip,keepaspectratio]{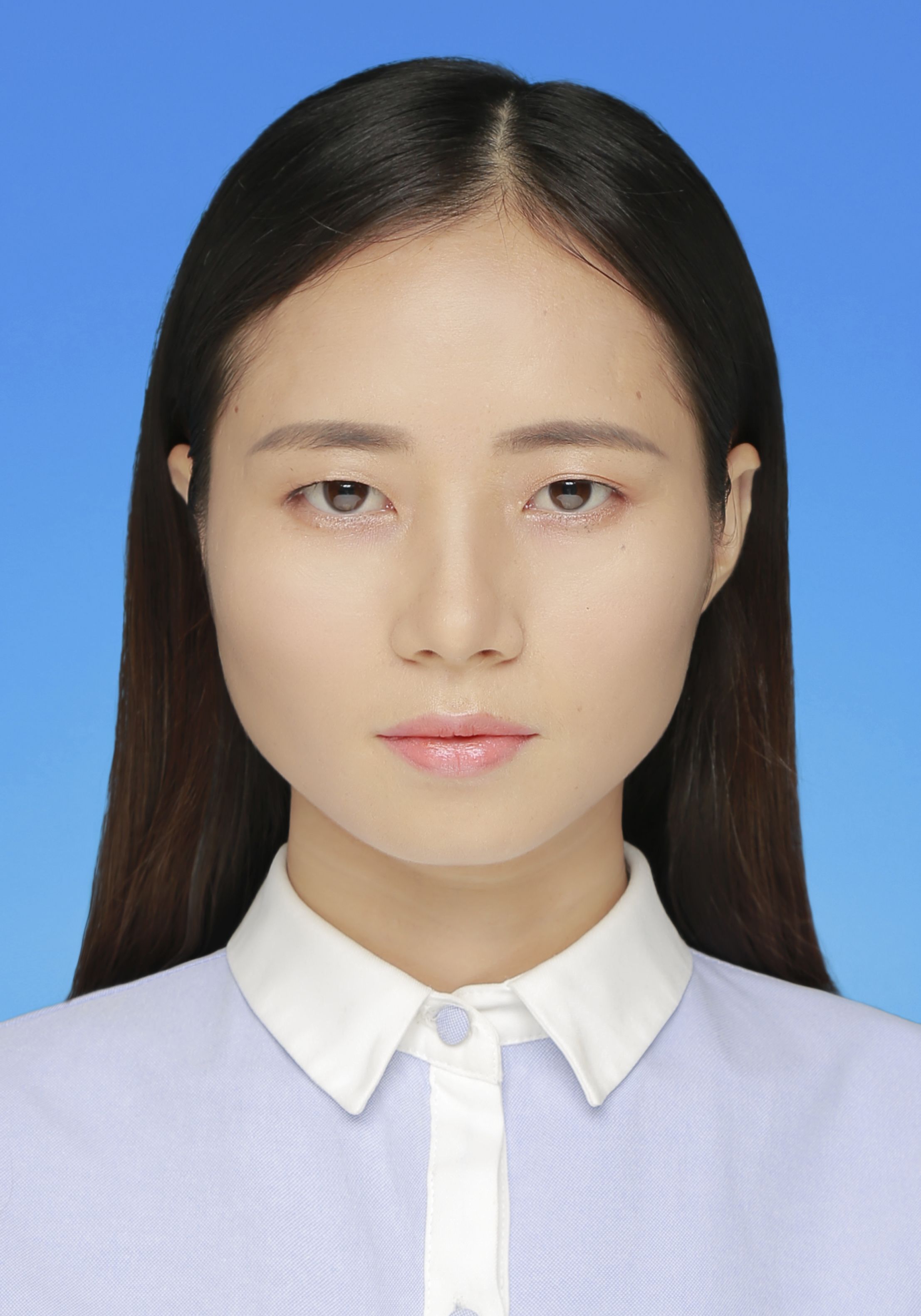}}]{Xiaoli Liu}
Xiaoli Liu received the Ph.D. degree from the Beijing University of Posts and Telecommunications, Beijing, China. She currently is a Postdoctoral Fellow with the School of Artificial Intelligence, Beijing University of Posts and Telecommunications, Beijing, China. Her research interests include computer vision, machine learning and image processing, and deep learning.
\end{IEEEbiography}

\begin{IEEEbiography}[{\includegraphics[width=1in,height=1.25in,clip,keepaspectratio]{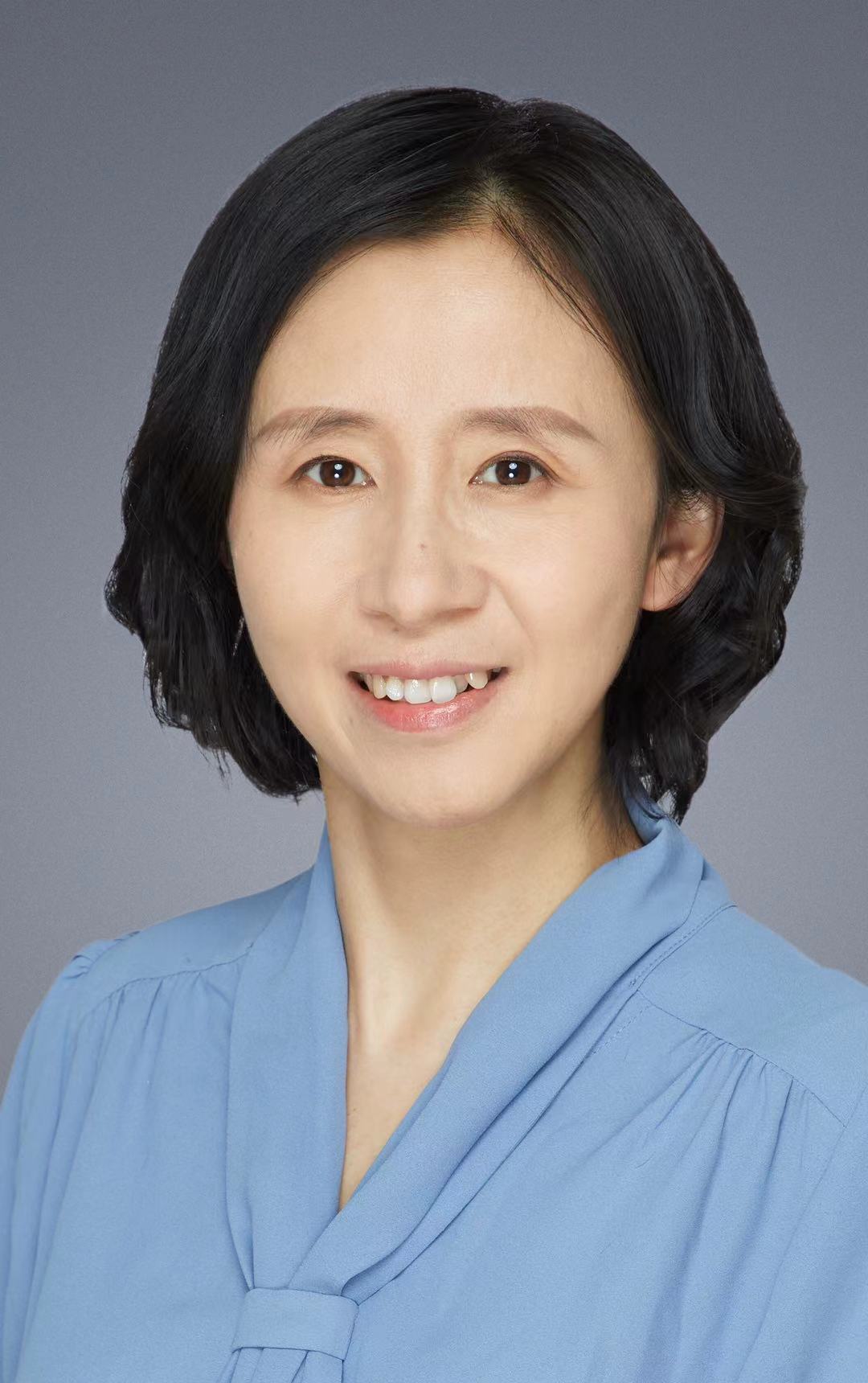}}]{Jianqin Yin}
Jianqin Yin received the Ph.D. degree from Shandong University, Jinan, China, in 2013. \\She currently is a Professor with the School of Artificial Intelligence, Beijing University of Posts and Telecommunications, Beijing, China. Her research interests include service robot, pattern recognition, machine learning and image processing. Email: jqyin@bupt.edu.cn.
\end{IEEEbiography}

\begin{IEEEbiography}[{\includegraphics[width=1in,height=1.25in,clip,keepaspectratio]{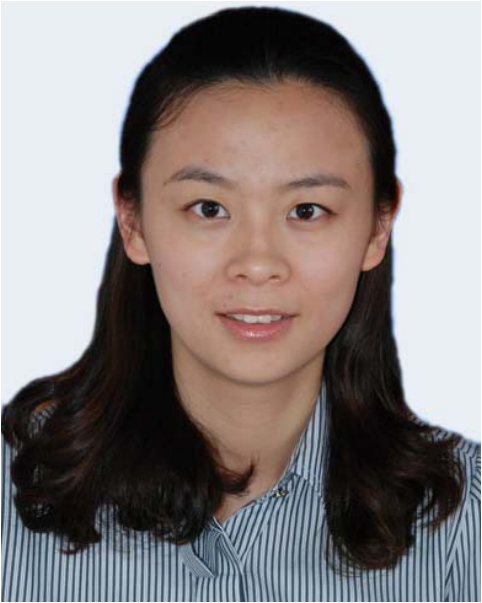}}]{Di Guo}
Di Guo received the Ph.D. degree in computer science and technology from Tsinghua University, Beijing, China, in 2017. She is currently a Professor with the School of Artificial Intelligence, Beijing University of Posts and Telecommunications, Beijing, China. Her research interests include intelligent robot, computer vision and machine learning.
\end{IEEEbiography}

\begin{IEEEbiography}[{\includegraphics[width=1in,height=1.25in,clip,keepaspectratio]{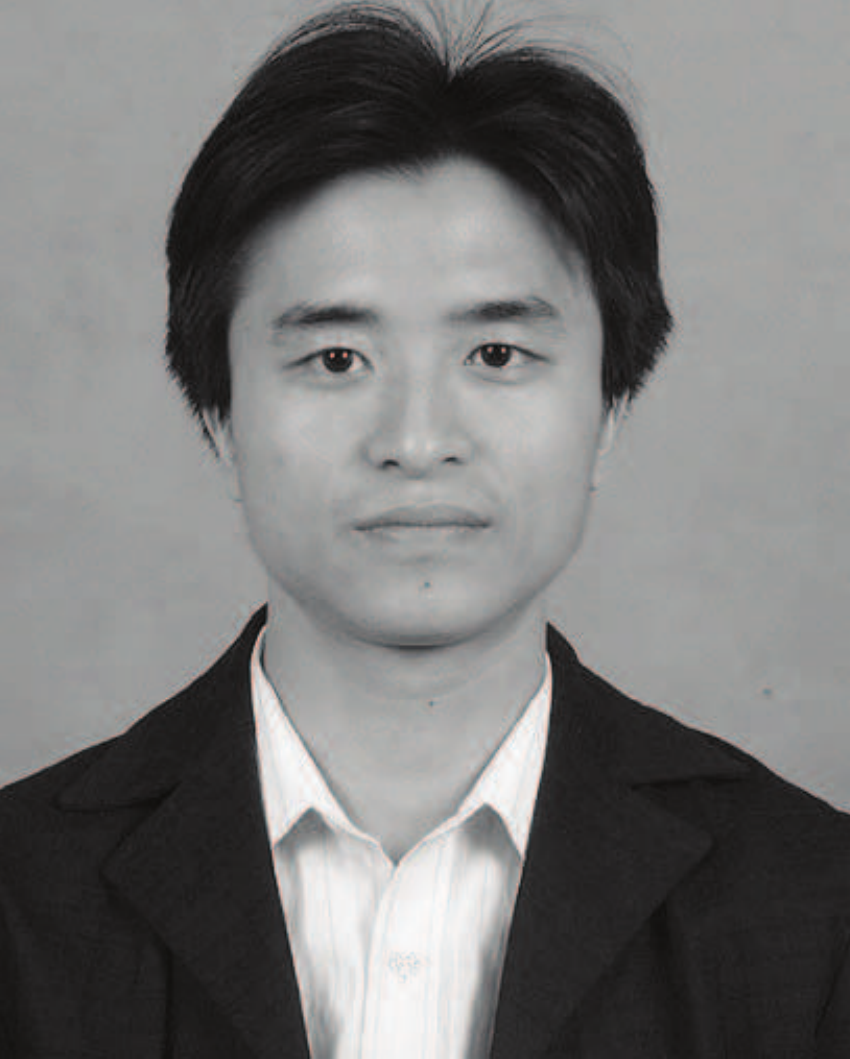}}]{Huaping Liu}
 Huaping Liu received the Ph.D. degree from Tsinghua University, Beijing, China, in 2004. \\He currently is an Associate Professor with the Department of Computer Science and Technology, Tsinghua University, Beijing, China. His research interests include robot perception and learning.\\ Dr. Liu serves as an Associate Editor of several journals including the IEEE ROBOTICS AND AUTOMATION LETTERS, Neurocomputing, Cognitive Computation, and some conferences including the International Conference on Robotics and Automation and the International Conference on Intelligent Robots and Systems. He also served as a Program Committee Member of RSS2016 and IJCAI2016.
 \end{IEEEbiography}

\end{document}